\documentclass[sigconf]{acmart}
\usepackage{subfigure}
\usepackage[noend]{algpseudocode}
\usepackage{algorithmicx,algorithm}

\usepackage{amssymb}
\usepackage{multirow}
\AtBeginDocument{%
  \providecommand\BibTeX{{%
    \normalfont B\kern-0.5em{\scshape i\kern-0.25em b}\kern-0.8em\TeX}}}

\copyrightyear{2022}
\acmYear{2022}
\setcopyright{acmcopyright}
\acmConference[MM '22]{Proceedings of the 30th ACM International Conference on Multimedia}{October 10--14, 2022}{Lisbon, Portugal}
\acmBooktitle{Proceedings of the 30th ACM International Conference on Multimedia (MM '22), October 10--14, 2022, Lisbon, Portugal}
\acmPrice{15.00}
\acmDOI{10.1145/3503161.3548040}
\acmISBN{978-1-4503-9203-7/22/10}

\begin{document}
\title[CST for Semi-supervised Object Detection with DCR]{Cycle Self-Training for Semi-Supervised Object Detection with Distribution Consistency Reweighting}



\author{Hao Liu}
\authornote{Both authors contributed equally to this research.}
\orcid{0000-0003-3659-5563}
\affiliation{
  \institution{Artificial Intelligence on Electric Power System State Grid Corporation Joint Laboratory(State Grid Smart Grid Research Institute Co., Ltd.)}
  \streetaddress{}
  \city{Beijing}
  \state{}
  \country{China}
  \postcode{}
}
\email{liuhao2018@ict.ac.cn}

\author{Bin Chen\authornotemark{*}}
\affiliation{
  \institution{Institute of Computing Technology, Chinese Academy of Sciences \& University of Chinese Academy of Sciences}
  \streetaddress{}
  \city{Beijing}
  \state{}
  \country{China}
  \postcode{}
}
\email{chenbin20s@ict.ac.cn}

\author{Bo Wang}
\affiliation{
  \institution{Artificial Intelligence on Electric Power System State Grid Corporation Joint Laboratory(State Grid Smart Grid Research Institute Co., Ltd.)}
  \streetaddress{}
  \city{Beijing}
  \state{}
  \country{China}
  \postcode{}
}
\email{flish\_wang@sina.com}

\author{Chunpeng Wu}
\affiliation{
  \institution{Artificial Intelligence on Electric Power System State Grid Corporation Joint Laboratory(State Grid Smart Grid Research Institute Co., Ltd.)}
  \streetaddress{}
  \city{Beijing}
  \state{}
  \country{China}
  \postcode{}
}
\email{chunpeng.wu@alumni.duke.edu}

\author{Feng Dai}
\affiliation{
  \institution{Institute of Computing Technology, Chinese Academy of Sciences}
  \streetaddress{}
  \city{Beijing}
  \state{}
  \country{China}
  \postcode{}
}
\email{fdai@ict.ac.cn}

\author{Peng Wu}
\affiliation{
  \institution{Artificial Intelligence on Electric Power System State Grid Corporation Joint Laboratory(State Grid Smart Grid Research Institute Co., Ltd.)}
  \streetaddress{}
  \city{Beijing}
  \state{}
  \country{China}
  \postcode{}
}
\email{wup@geiri.sgcc.com.cn}
\renewcommand{\shortauthors}{Hao Liu et al.}

\begin{abstract}
Recently, many semi-supervised object detection (SSOD) methods adopt teacher-student framework and have achieved state-of-the-art results. However, the teacher network is tightly coupled with the student network since the teacher is an exponential moving average (EMA) of the student, which causes a performance bottleneck. To address the coupling problem, we propose a Cycle Self-Training (CST) framework for SSOD, which consists of two teachers T1 and T2, two students S1 and S2. Based on these networks, a cycle self-training mechanism is built, i.e., S1${\rightarrow}$T1${\rightarrow}$S2${\rightarrow}$T2${\rightarrow}$S1. For S${\rightarrow}$T, we also utilize the EMA weights of the students to update the teachers. For T${\rightarrow}$S, instead of providing supervision for its own student S1(S2) directly, the teacher T1(T2) generates pseudo-labels for the student S2(S1), which looses the coupling effect. Moreover, owing to the property of EMA, the teacher is most likely to accumulate the biases from the student and make the mistakes irreversible. To mitigate the problem, we also propose a distribution consistency reweighting strategy, where pseudo-labels are reweighted based on distribution consistency across the teachers T1 and T2. With the strategy, the two students S2 and S1 can be trained robustly with noisy pseudo labels to avoid confirmation biases. Extensive experiments prove the superiority of CST by consistently improving the AP over the baseline and outperforming state-of-the-art methods by 2.1\% absolute AP improvements with scarce labeled data.
\end{abstract}




\maketitle

\begin{figure}
	\centering
	\subfigure[]{
		\begin{minipage}[b]{0.22\textwidth}
				\includegraphics[width=\textwidth]{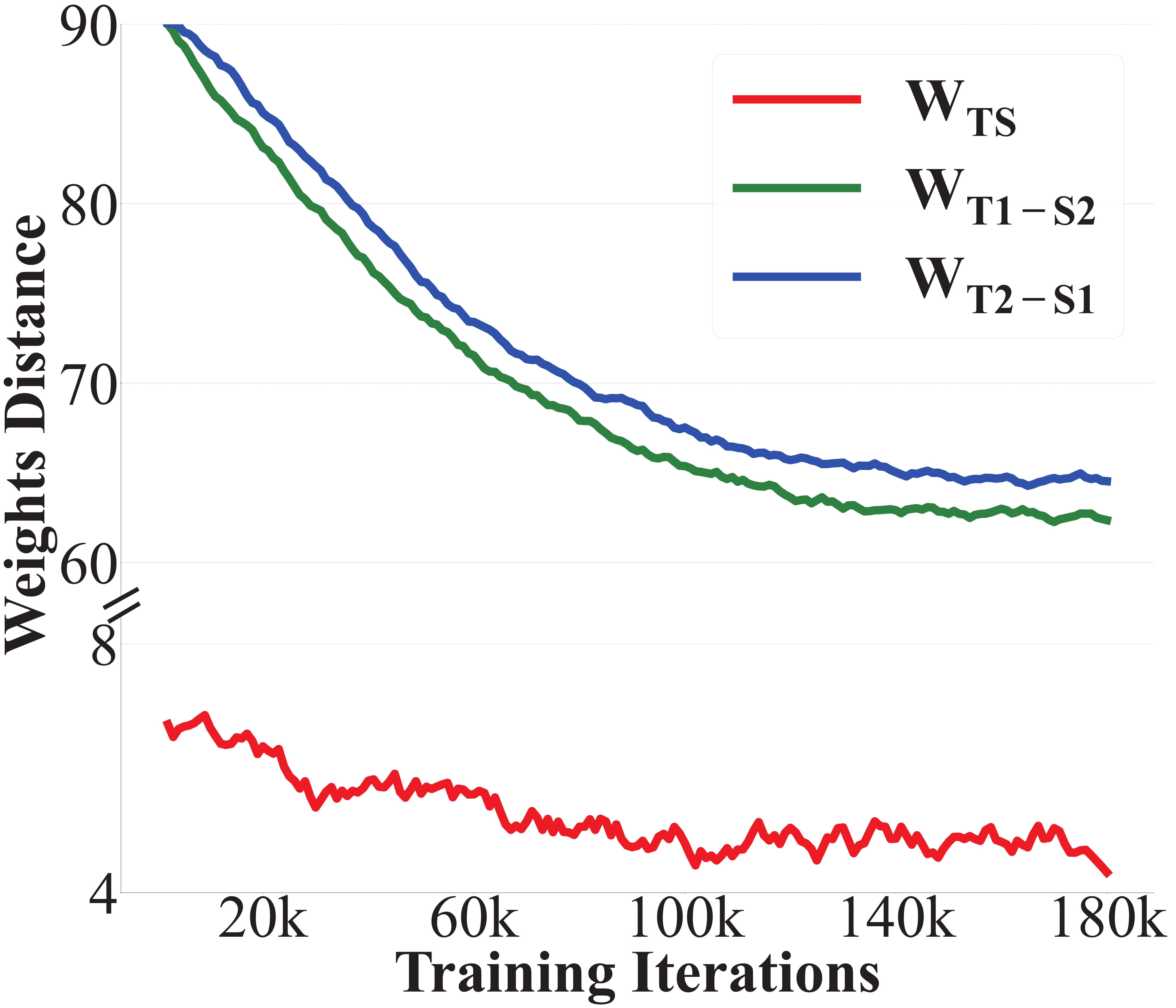}
		\end{minipage}
	}
	\subfigure[]{
		\begin{minipage}[b]{0.22\textwidth}
				\includegraphics[width=\textwidth]{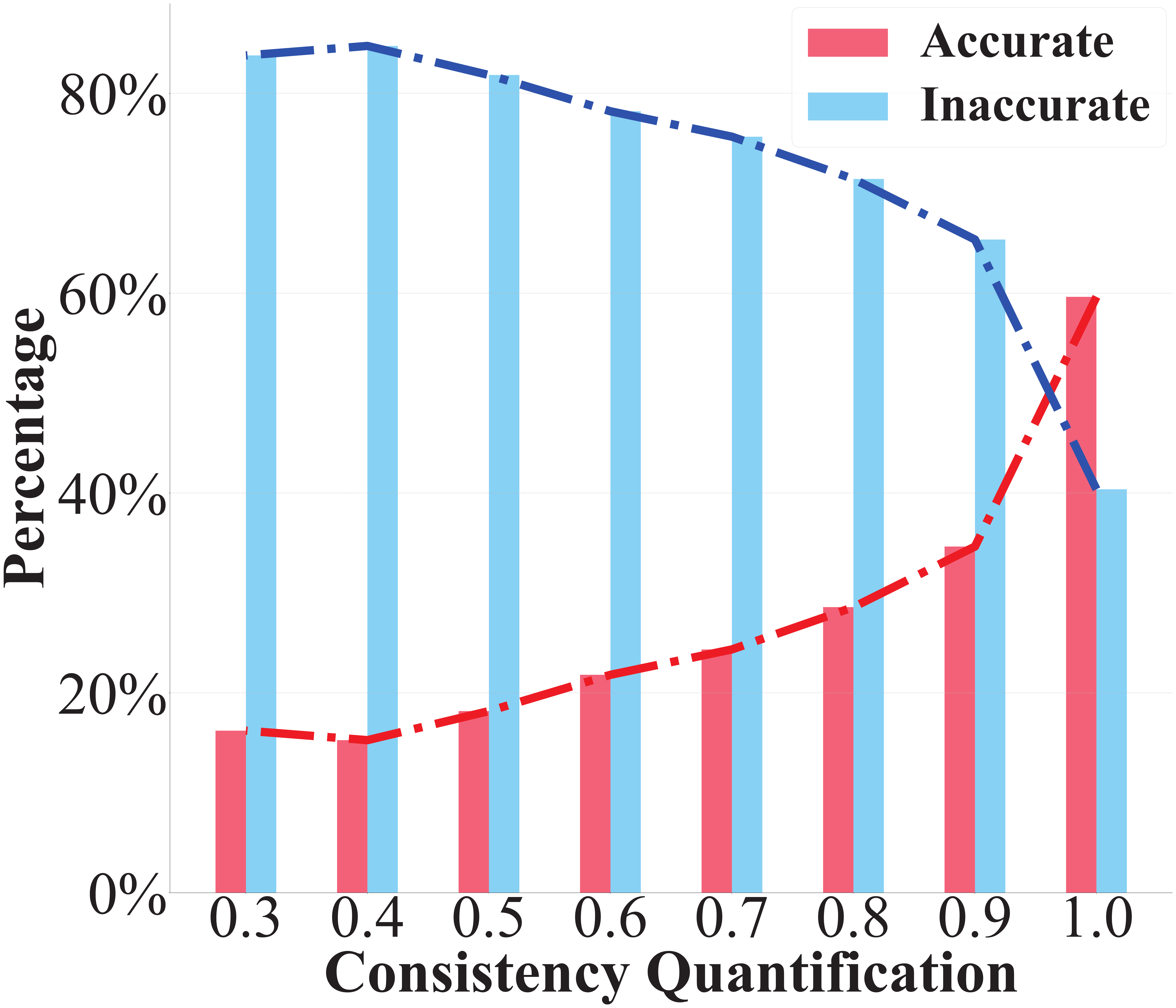}
		\end{minipage}
	}
	\subfigure[]{
		\begin{minipage}[b]{0.22\textwidth}
				\includegraphics[width=\textwidth]{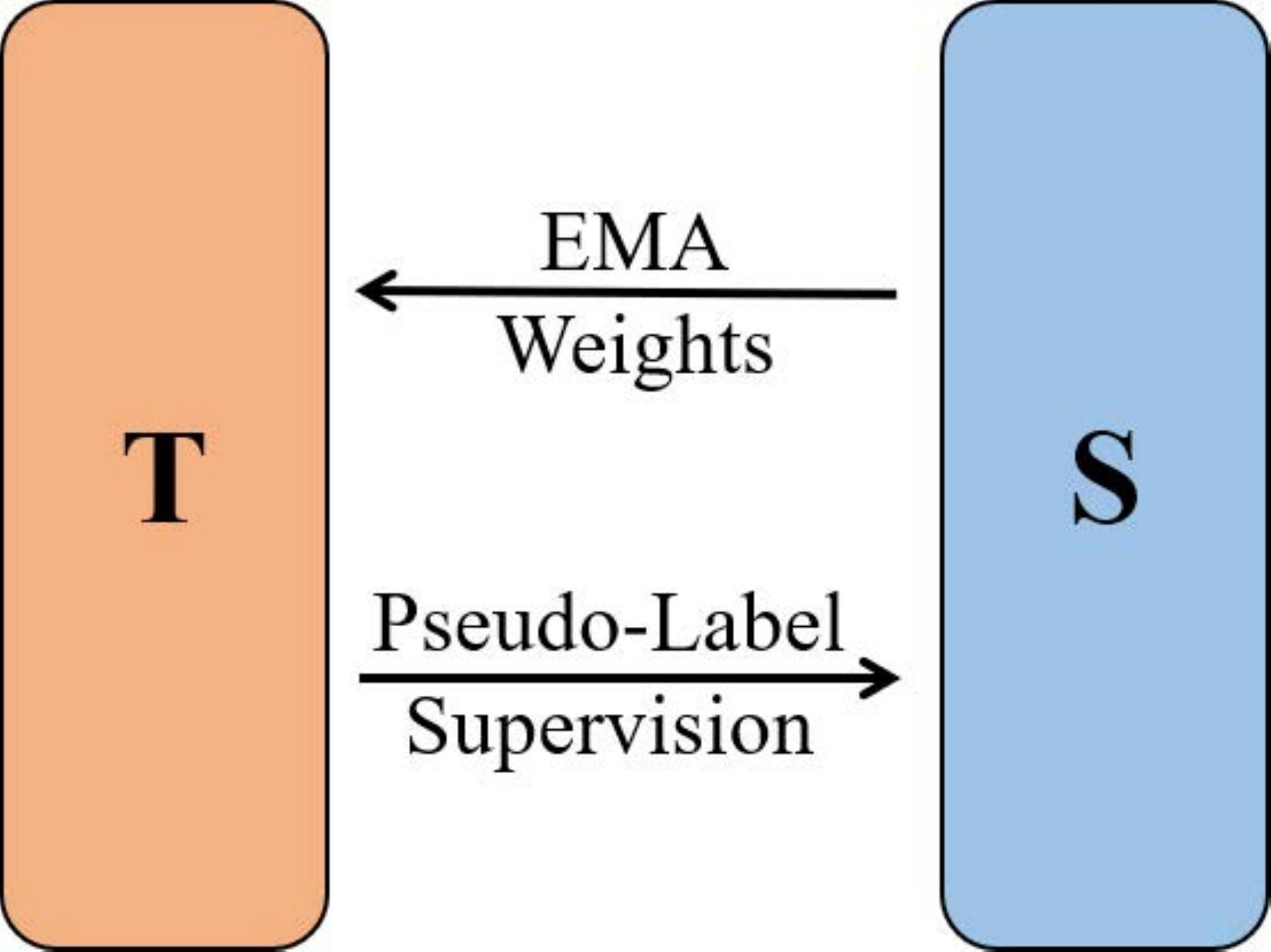}
		\end{minipage}
	}
	\subfigure[]{
		\begin{minipage}[b]{0.22\textwidth}
				\includegraphics[width=\textwidth]{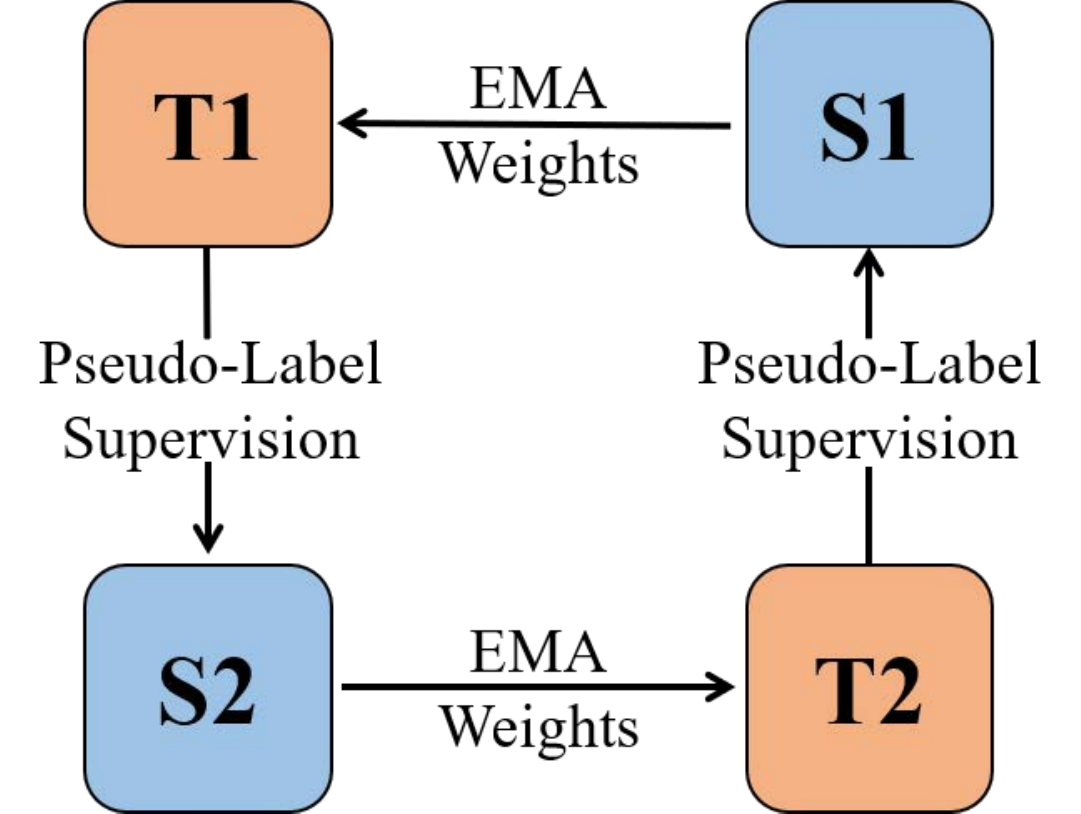}
		\end{minipage}
	}
	\caption{Teacher-Student Framework versus Cycle Self-Training Framework. (a) Euclidean distance of weights. (b) Percentage of accurate/inaccurate samples. (c) Teacher-Student Framework. (d) Cycle Self-Training Framework.}
  \label{fig:intro}
\end{figure}

\section{Introduction}
With the rapid development of deep learning, large amount of labeled data become the critical component
during training process. 
However, collecting labels is time-consuming and expensive \cite{openimage}, especially
for object detection with instance-level annotations. 
This has encouraged Semi-Supervised Learning (SSL)
methods to leverage unlabeled data, such as image classification \cite{shi2018transductive,xie2020self,lim2021class,assran2021semi,hu2021simple} and object detection \cite{sohn2020simple,zhou2021instant,yang2021interactive,tang2021humble,liu2020unbiased} 
tasks. This paper studies the problem of semi-supervised object detection (SSOD) that focuses not only on classification
but also localization.

For SSOD, recent mainstream methods are based on pseudo-labeling, which train detectors on labeled data and 
unlabeled data with pseudo labels jointly to improve detection performance. The pseudo-labeling based
model consists of two components: a teacher model and a student model, as shown in Figure \ref{fig:intro} (c). The teacher generates pseudo-labels
to train the student, and the student updates the knowledge it learned back to the teacher. Also, the teacher
can be regarded as the temporal ensemble of the student \cite{liu2020unbiased}, which generates more stable and accurate pseudo-labels.
However, the teacher is tightly coupled with the student due to the exponential moving average (EMA). As shown
in Figure \ref{fig:intro} (a), the Euclidean distance of weights $W_{TS}$ between the EMA teacher and student is very close, and their 
distance continues to drop during training process. It demonstrates that the teacher does not have more
meaningful knowledge compared to the student, which limits the descriptive ability of the existing Teacher-Student
framework. Moreover, due to the inherent nature of imbalance in object detection tasks and datasets \cite{Imbalance},
the teacher is prone to generate biased pseudo-labels towards dominant classes, making the imbalanced
problem even more severe. However, the teacher is likely to accumulate the biases
from the student and make the mistakes irreversible on account of the property of EMA, which is a case of the confirmation bias \cite{tarvainen2017mean}. 

To address the tightly coupling problem, we introduce a Cycle Self-Training (CST) framework for 
semi-supervised object detection. The proposed CST framework consists of four sub-networks: two teacher
networks T1 and T2, two student networks S1 and S2, as shown in Figure \ref{fig:intro} (d). To overcome the coupling effect of the Teacher-Student framework, we propose to loose the constraint between T1 and S1, i.e., T1 network does not
provide pseudo-labels any longer for S1 network directly. Instead, a cycle self-training framework is built 
based on the following four relationships: (1) the knowledge that the student network S1 learned is transferred
to the teacher network T1 via EMA; (2) the teacher network T1 provides supervision for the student network
S2; (3) the EMA of the student network S2 is used to update the teacher network T2; (4) the teacher network T2
generates pseudo-labels for the student network S1. With the proposed CST framework, we construct a knowledge
transferring loop so that the student networks S1 and S2 can acquire more meaningful knowledge.
As shown in Figure \ref{fig:intro} (a), compared with the typical Teacher-Student framework, the weights distances $W_{T1-S2}$ and $W_{T2-S1}$ between the teacher
network T1 (T2) and S2 (S1) always keep larger from each other, which demonstrates that the teacher and the student
networks in our CST framework are loosely coupled.

To tackle the problem of confirmation bias, we propose a distribution consistency reweighting strategy, where
pseudo-labels are learned conditioned on distribution consistency. If the predicted classification distributions of a pseudo box 
across the teacher networks T1 and T2 are sufficiently consistent, we consider it as a stable instance and increase 
its influence in the training process via consistency reweighting. As shown in Figure \ref{fig:intro} (b), 
according to our statistics, the accuracy of pseudo labels and the consistency quantification 
illustrate a strong positive correlation, which means that the pseudo labels with high consistency 
values can provide more accurate category information. With our proposed strategy, the two student 
networks S1 and S2 can be trained robustly with noisy pseudo labels to avoid accumulating confirmation 
biases.

In summary, our main contributions can be summarized as: 
\begin{itemize}
\item We propose a Cycle Self-Training (CST) framework for semi-supervised object detection, in which
a knowledge transferring loop is built to loose the tightly coupling effect of the teacher-student framework.
\item A distribution consistency reweighting strategy is proposed to incorporate with our CST framework so that the two student networks can be trained robustly with noisy pseudo labels to avoid the accumulating confirmation biases. 
\item We conduct extensive experiments on COCO \cite{lin2014microsoft} and PASCAL VOC \cite{everingham2010pascal} datasets, which validates the effectiveness of our proposed framework. Specifically, the CST framework obtains consistent improvements over the baseline and outperforms the state-of-the-art methods by 2.1\% absolute AP improvements with scarce labeled data.
\end{itemize}

\section{Related Work}

\subsection{Object Detection}

Object detection tries to determine the category and location of each object instance appearing in an image. 
Nowadays, numerous methods to deal with this problem can be roughly divided into two main pipelines: two-stage and one-stage. The two-stage pipeline first 
uses a Region Proposal Network (RPN) to generate some proposals, which are coarsely localized and categorized 
into the foreground or not, and then refines the proposals by multi-classification and 
further regression \cite{FasterRCNN,dai2016r,cai2018cascade,chen2019hybrid,song2020revisiting}. 
The one-stage pipeline omits the region proposals generation process and directly gives the classification and 
localization results of the anchor boxes \cite{YOLO,SSD,lin2017focal}. 
Meanwhile, some one-stage methods formulate bounding box object detection as detecting paired or triplet key-points 
\cite{law2018cornernet,tian2019fcos,duan2019centernet,dong2020centripetalnet}. 
Moreover, recent works present a paradigm 
based on the Transformer \cite{vaswani2017attention} structure to detect objects \cite{carion2020end,zhu2020deformable,liu2021swin}, 
which achieve the state-of-the-art performance on the current object detection datasets. 
However, both of these works give poor scalability due to the limitation of the quantity of labeled data.

\begin{figure*}
	\centering
	\includegraphics[width=0.94\textwidth]{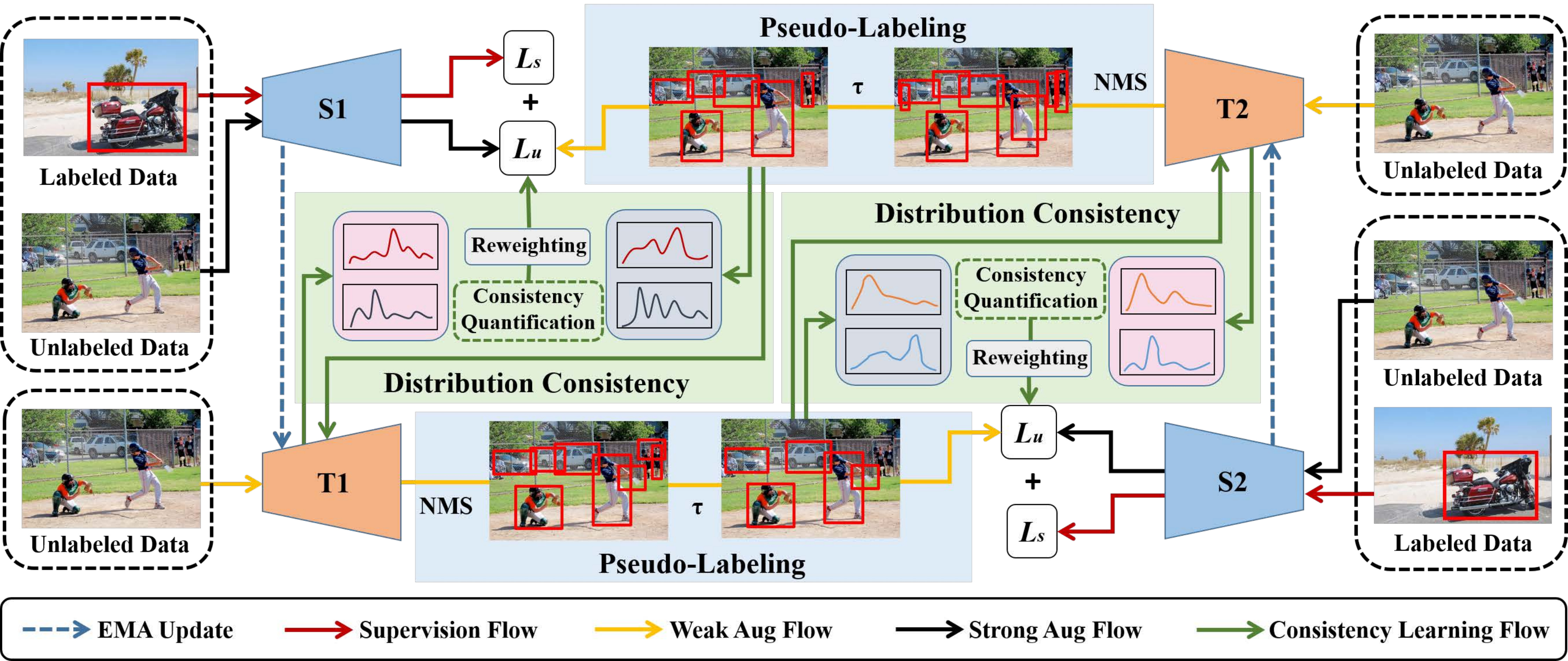}
	\caption{The overview of the cycle self-training (CST) framework with distribution consistency reweighting (DCR) strategy. Labeled and unlabeled images form the training data batch. In each iteration, the teacher T1 (T2) perform pseudo-labeling on weak augmented images to train the student S2 (S1) with strong augmented images. And the student S1 (S2) is utilized to update the teacher T1 (T2) via EMA. Moreover, consistency quantification is performed across the two teachers to reweight unsupervised loss. The final loss is the sum of supervised loss ${L_s}$ and unsupervised loss ${L_u}$.}
  \label{fig:structure}
\end{figure*}

\subsection{Semi-supervised Learning (SSL)}

In recent years, semi-supervised learning (SSL) has made some great progress by 
mining the potential of the unlabeled data, which is beneficial for 
the tasks with limited labeled data and abundant unlabeled data. 
Existing semi-supervised learning approaches are based on the consistency regularization principle or pseudo labeling technique. 
Dual Student \cite{ke2019dual} explains the coupling effect of the EMA and replaces the teacher with another student to address this problem. 
RemixMatch \cite{berthelot2019remixmatch} produces distribution alignment and augmentation anchoring to reduce 
the quantity labeled data. 
UDA \cite{xie2020unsupervised} substitutes the traditional noise injection with the 
high quality data augmentation to improve consistency semi-supervised learning. 
Fixmatch \cite{sohn2020fixmatch} applies the consistency regularization and introduces artificial labels on weakly augmented unlabeled images. 
All of these methods leverage the augmentations or perturbations applied to original input images 
to ensure consistent output predictions. 
The other kind of methods utilize the pseudo labels by adopting the teacher-student co-training framework. 
Deep co-training \cite{peng2020deep} presents a deep adversarial co-training approach and explores the effect on the  
prediction diversity of the teacher-student models. Noisy Student \cite{xie2020self} uses unlabeled data 
by adding additional noise to the student for better learning the knowledge of the teacher model.
FlexMatch \cite{zhang2021flexmatch} proposes Curriculum Pseudo Labeling to dynamically leverage unlabeled data with thresholds.
SemiMatch \cite{kim2022semi} generates pseudo-labels between source and weakly augmented target 
to learn the model again between source and strongly augmented one.

\subsection{Semi-supervised Object Detection (SSOD)}

To avoid the large cost restriction of detection annotations, currently, the 
semi-supervised learning is applied to object detection task. CSD \cite{jeong2019consistency} directly uses the simple flip augmentation 
to input images based on consistency regularization principle, and the loss function is built on 
the consistency of the two output predictions. STAC \cite{sohn2020simple} designs a special framework 
based on both pseudo labels technique and consistency regularization for object detection, 
including a fixed teacher network for pseudo labels generation and a student network for training with strong augmentations. 
Unbiased Teacher \cite{liu2020unbiased} further explores the teacher-student framework and updates the weights of the teacher network by the EMA technique,  
which uses Focal Loss \cite{lin2017focal} to mitigate the class-imbalance issue to some degree. Thus it is also treated as the comparison baseline in most cases. 
Instant-Teaching \cite{zhou2021instant} leverages Mixup \cite{zhang2018mixup} and Mosaic \cite{bochkovskiy2020yolov4} augmentations and 
proposes a co-rectify scheme to alleviate the confirmation bias \cite{tarvainen2017mean}. 
Humble Teacher \cite{tang2021humble} utilizes plenty of region proposals and soft pseudo labels for training with a light-weighted 
detection-specific data ensemble algorithm. ISMT \cite{yang2021interactive} proposes an interactive self-training framework 
and performs NMS operation to fuse the results of the current and historical iteration to improve the quality of pseudo labels. 
Soft Teacher \cite{xu2021end} proposes a mechanism where the classification loss is weighted by the score generated from teacher
and a box jittering strategy to select reliable pseudo boxes for regression. 
Combating Noise \cite{wang2021combating} treats uncertainty quantification as the soft target and facilitates multi-peak
probability distribution. 
CPL \cite{li2021rethinking} introduces certainty-aware pseudo labels and uses dynamic thresholds to mitigate the class imbalance problem.
MUM \cite{kim2021mum} proposes the Mix and UnMix data augmentation method to generate strongly-augmented images for training, 
which can be easily equipped on other SSOD methods. 

\section{Methodology}


In this paper, we propose a cycle self-training framework with distribution consistency reweighting strategy to overcome 
the coupling and the confirmation biases problems. 
The overall structure of our framework is shown in Figure \ref{fig:structure}. 

\subsection{Preliminary}
For semi-supervised object detection (SSOD), a set of labeled data ${D_l=\{(x_i^l,y_i^l)\}_{i=1}^{N_l}}$
and unlabeled data ${D_u=\{x_i^u\}_{i=1}^{N_u}}$ are available for training, where ${x}$ and ${y}$ denote image
and ground-truth annotations respectively, i.e., class labels and bounding box coordinates, ${N_l}$ and
${N_u}$ represent the number of labeled and unlabeled data. And the ultimate goal of SSOD is to improve
detection accuracy by training object detectors on both labeled and unlabeled data.

To leverage the unlabeled images, we also adopt the Teacher-Student training paradigm, where the Student
is optimized by using the pseudo-labels generated from the Teacher, and the Teacher is updated by
gradually transferring the weights of continually learned Student model, similar to recent works \cite{zhou2021instant,yang2021interactive,tang2021humble,wang2021combating,xu2021end,liu2020unbiased,kim2021mum}.
Also, a confidence threshold of predicted bounding boxes is set to filter low-confidence predicted
bounding boxes, which are more likely to be false positive samples. Moreover, to address the duplicated
boxes prediction issue existing in object detection, we eliminate redundancy by applying non-maximum 
suppression (NMS) before confidence filtering with threshold $\tau$. 
And different augmentation strategies are adopted 
for teacher and student model respectively, i.e., weak augmentation for pseudo-labeling of teacher
model and strong augmentation for training of student model. 

After obtaining the pseudo-labels of unlabeled images from the teacher model, a mixed batch of equal
numbers of labeled and unlabeled images are randomly sampled to feed into the supervised branch and
unsupervised branch respectively. The final loss ${L}$ is the weighted sum of the supervised loss and 
unsupervised loss,
\begin{equation}
L=L_s+\alpha L_u,
\end{equation}
where ${L_s}$ and ${L_u}$ denote the supervised loss of labeled images and unsupervised loss of unlabeled
images respectively, ${\alpha}$ is the weight factor to balance these two losses. And the supervised loss
and the unsupervised loss are defined as follows:
\begin{equation}
L_s=\frac{1}{n_l}\sum_{i=1}^{n_l}(L_{cls}(x_i^l, y_i^l)+L_{reg}(x_i^l, y_i^l)),
\end{equation}
\begin{equation}
L_u=\frac{1}{n_u}\sum_{i=1}^{n_u}L_{cls}(x_i^u, \hat{y}_i^u),
\end{equation}
where ${L_{cls}}$ is the classification loss, ${L_{reg}}$ is the regression loss, ${n_l}$ and ${n_u}$ denote
the number of labeled images and unlabeled images in a training batch, ${x_i^l}$ and ${y_i^l}$ indicate the 
${i}$-th labeled image and corresponding ground-truth label, while ${x_i^u}$ and ${\hat{y}_i^l}$ denote the
${i}$-th unlabeled image and its pseudo-label from teacher model. Similar to \cite{liu2020unbiased}, we also do not apply
regression loss for unlabeled images since predicted confidence can not show the localization quality.

Besides, our proposed self-training framework can be applied to mainstream object detectors, including two-stage 
detectors \cite{FasterRCNN,dai2016r,cai2018cascade,chen2019hybrid,song2020revisiting} and one-stage 
detectors \cite{YOLO,SSD,lin2017focal,law2018cornernet,tian2019fcos}. 
For fair comparisons, we also utilize
Faster R-CNN \cite{FasterRCNN} to clarify our method.

\subsection{Cycle Self-Training Framework}
\label{sec:cst}

Our proposed Cycle Self-Training framework consists of three stages, the Burn-In stage, the training stage and
the inference stage. 

\textbf{Burn-In Stage.} In the training stage, the student network is supervised by the pseudo-labels generated
from the teacher network. Hence, the quality of pseudo-labels is very important for detection performance. So it
is necessary to have a good initialization for both student and teacher networks \cite{liu2020unbiased}. In our framework, the students
S1 and S2 are initialized with different parameters firstly. Then we only utilize the available labeled images
to optimize model ${\theta_1}$ (${\theta_2}$) with the supervised loss ${L_s}$ for a fixed amount of iterations. 
After the Burn-In stage, the weights ${\theta_1}$ (${\theta_2}$) is copied to both the teacher T1 (T2) and the student S1 (S2), i.e., 
${\theta_1}$ ${\rightarrow}$ ${\theta_1^t}$, ${\theta_1}$ ${\rightarrow}$ ${\theta_1^s}$ (${\theta_2}$ ${\rightarrow}$ ${\theta_2^t}$,
${\theta_2}$ ${\rightarrow}$ ${\theta_2^s}$). Based on the initialized parameters, the models are further trained 
with the proposed cycle self-training mechanism to improve performance.

\textbf{Training Stage.} As analyzed in the previous section, due to the coupling effect, the typical EMA teacher can not 
provide more meaningful knowledge for student along with the training process, especially in the later 
stage. To address the issue, an intuitive idea is to loose the couping between the teacher and the student.
Hence, we propose a cycle self-training (CST) framework for semi-supervised object detection, in which 
a knowledge transferring loop is built to loose the coupling effect of the Teacher-Student framework.

As shown in Figure \ref{fig:structure}, the proposed CST framework consists of four sub-networks: two teacher networks T1 
and T2, and two student networks S1 and S2. To eliminate the coupling effect between T1 and S1, a cycle
self-training mechanism is built, i.e., S1 ${\rightarrow}$ T1 ${\rightarrow}$ S2 ${\rightarrow}$ T2 ${\rightarrow}$ S1
among the four networks. For S ${\rightarrow}$ T, we also update the teacher networks with the EMA weights of
the student networks. By doing so, the advantage of the typical Teacher-Student framework is preserved, i.e.,
the imbalanced pseudo-labeling biased issue can be alleviated and more stable pseudo-labels can be obtained \cite{liu2020unbiased}.
For T ${\rightarrow}$ S, instead of providing supervision for its own student S1 (S2) directly, the teacher T1 (T2)
generates pseudo-labels for the student S2 (S1), which looses the coupling relationship and transfers more 
meaningful knowledge indirectly.

Besides, due to the existence of EMA relationship, the EMA teacher still accumulates the mistakes from its own
student and enforces the other student to follow. To overcome this problem, we propose an consistency
learning strategy to avoid accumulating biases and collapsing into each other. Specifically, as shown in Figure \ref{fig:structure}, 
the pseudo-labels generated from the teacher T1 are measured by the teacher T2 in Distribution Consistency module, which is to perform consistency
quantification of classification distribution. Based on the consistency values, each pseudo-label proposal 
is re-weighted to train the student S2 robustly. More details about the learning strategy will be described in the 
next section.

\textbf{Inference Stage.} At the stage of inference, only the teacher networks are utilized. To illustrate the effect of our framework, we test the single teacher for fair comparisons with the recent state-of-the-art methods. Besides, the teachers T1 and T2 own different parameters, which perform differently for the same categories. So we also report the ensemble results of these two teachers with the Weighted Box Fusion (WBF) \cite{solovyev2021weighted} method, denoted as CST${^*}$.

\subsection{Distribution Consistency Reweighting}
\label{sec:dcr}

\begin{figure}
	\centering
	\includegraphics[width=0.47\textwidth]{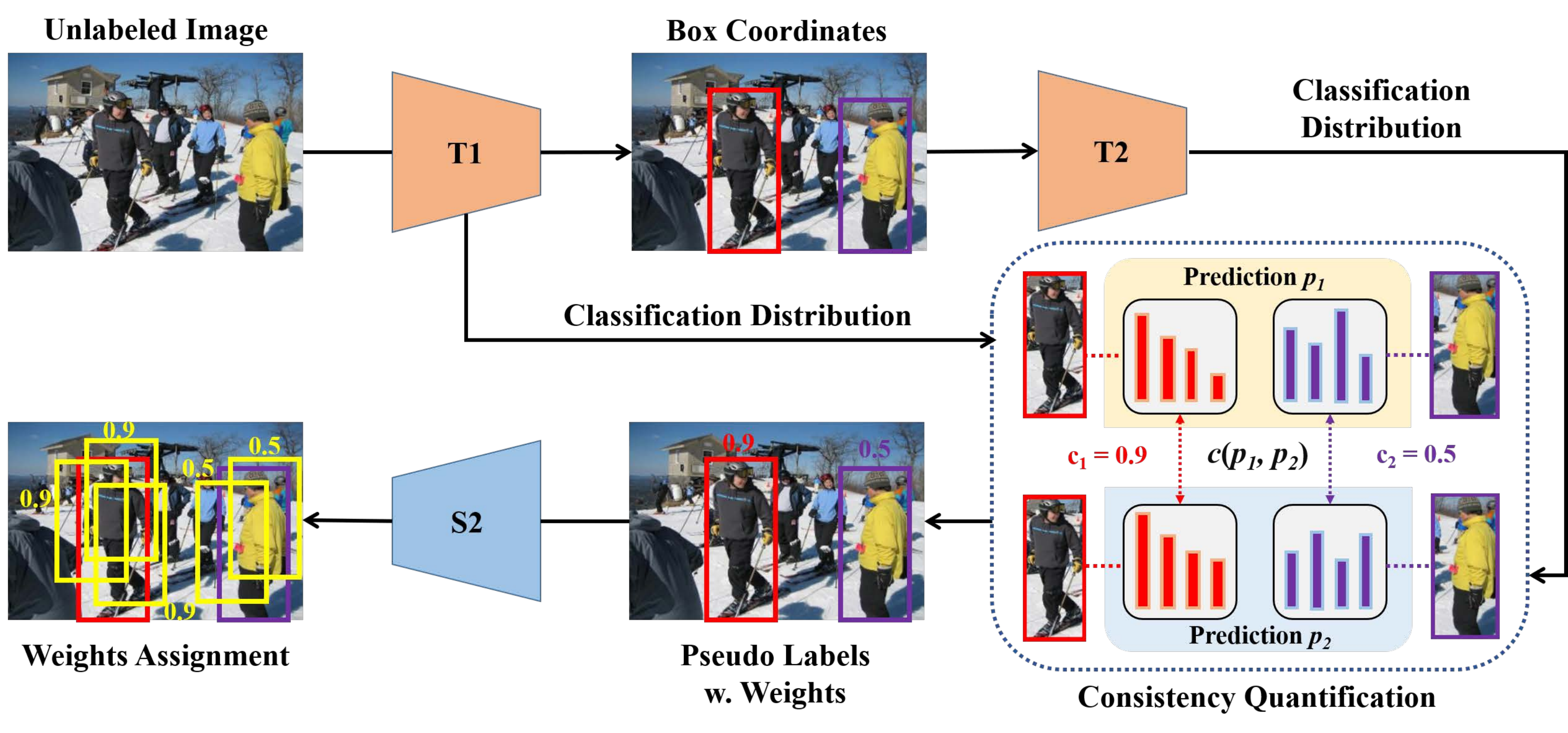}
	\caption{The structure of distribution consistency reweighting (DCR). Pseudo-labels generated by the teacher T1 are measured by the teacher T2 to perform consistency quantification $c(p_1, p_2)$ of classification distribution, and subsequently acts on the weights assignment for the student S2.}
  \label{fig:qua}
\end{figure}

As analyzed above, although the cycle self-training framework looses the coupling effect, there still exists confirmation bias issue due to EMA. To handle the above noise, a distribution consistency reweighting strategy 
is proposed, where pseudo-labels are learned conditioned on consistency, i.e., the prediction consistency 
between the teacher T1 and the teacher T2. If the predicted classification distributions of a pseudo box 
across the two teachers are consistent, we consider it as a stable and well-learned bounding box and increase 
its influence in the training process. By doing this, the EMA teacher T1 (T2) can not enforce the student S2 (S1) 
to be consistent with all pseudo-labels to avoid accumulating biases.

\textbf{Consistency Quantification.} As shown in Figure \ref{fig:qua}, the EMA teacher T1 generates pseudo-labels for the 
unlabeled image, including the box coordinates ${\{b_i\}}$ and corresponding classification 
distributions ${\{p_1^i\}}$. Then the EMA teacher T2 takes the pseudo boxes as input and predicts 
classification distribution for each box through its detection head, denoted as ${\{p_2^i\}}$. 
Given the predicted distributions from T1 and T2, we can perform consistency quantification by calculating 
the differences between distributions to assess the quality of all pseudo boxes generated by T1. 
For comparison, we explore two kinds of quantification styles. And the results are shown in the ablation 
studies.

\begin{itemize}
\item The L1 distance between the predicted distributions from T1 and T2. To make the quantification values more suitable for loss reweighting, we firstly normalize the L1 distance to ${0.5\sim1}$ with a \textit{sigmoid} mapping function. Then a linear normalization is performed so that the quantification values are in the range ${0\sim1}$. The specific formula is defined as follows:
	\begin{equation}
	  c_i(p_1^i, p_2^i) =  2\times(1-Sigmoid(||p_1^i-p_2^i||_{1}))
	\end{equation}
	where ${c_i}$ denotes the quantification value of the ${i}$th pseudo box, ${p_1^i}$ and ${p_2^i}$ are the corresponding distributions from T1 and T2 respectively. 
\item The JS divergence between the predicted distributions from T1 and T2. Because the JS values are also in the range ${0\sim1}$, we only utilize a tunable focusing parameter ${\beta=2}$ so that inconsistent instances are further down-weighted. The specific formula is expressed as follows:
	\begin{equation}
		c_i(p_1^i, p_2^i) =  JS(p_1^i, p_2^i)^\beta
	\end{equation}
\end{itemize}

\begin{algorithm}[!t]
\caption{Training procedure of the proposed CST}
\label{alg} 
	\begin{flushleft}
		\textbf{Require:} ${(X^l,Y^l),\ X^u}$: pair of labeled images and its annotations, and unlabeled images \\
		\textbf{Require:} ${f_{T1}(\cdot),\ f_{T2}(\cdot)}$: teacher object detection model T1 and model T2 \\
		\textbf{Require:} ${f_{S1}(\cdot),\ f_{S2}(\cdot)}$: student object detection model S1 and model S2 \\
		\textbf{Require:} ${w(\cdot),\ s(\cdot)}$: weak and strong augmentation \\
		\textbf{Require:} ${c(\cdot)}$: consistency quantification function \\
		\textbf{Require:} ${h(\cdot),\ \alpha}$: loss function and balancing weight \\
		\begin{algorithmic}[1]
			\For{each iter ${\in}$ [1, max\_iterations]} \\
				\quad\textbf{Prepare Data} \\
					\qquad${D\leftarrow w(X^l)+s(X^l),\ W\leftarrow w(X^u),\ S\leftarrow s(X^u)}$ \\
				\quad\textbf{Generate Pseudo Labels} \\
					\qquad${\hat{Y}^u_{S2} \leftarrow f_{T1}(W),\ \hat{Y}^u_{S1} \leftarrow f_{T2}(W)}$ \\
				\quad\textbf{Distribution Consistency Quantification} \\
					\qquad${C_{S2}\leftarrow c(\hat{Y}^u_{S2}, f_{T2}(\hat{Y}^u_{S2})),\ C_{S1}\leftarrow c(\hat{Y}^u_{S1}, f_{T1}(\hat{Y}^u_{S1}))}$ \\
				\quad\textbf{Compute the Supervised Loss} \\
					\qquad${P^l_{S1}\leftarrow f_{S1}(D),\ P^l_{S2}\leftarrow f_{S2}(D)}$ \\
					\qquad${L^l_{S1}\leftarrow h(P^l_{S1}, Y^l),\ L^l_{S2}\leftarrow h(P^l_{S2}, Y^l)}$ \\
				\quad\textbf{Compute the Unsupervised Loss} \\
					\qquad${P^u_{S1}\leftarrow f_{S1}(S),\ P^u_{S2}\leftarrow f_{S2}(S)}$ \\
					\qquad${L^u_{S1}\leftarrow h(P^u_{S1}, \hat{Y}^u_{S1}, C_{S1}),\ L^u_{S2}\leftarrow h(P^u_{S2}, \hat{Y}^u_{S2}, C_{S2})}$ \\
				\quad\textbf{Compute the Total Loss} \\
					\qquad${L_{S1}\leftarrow L^l_{S1}+\alpha L^u_{S1},\ L_{S2}\leftarrow L^l_{S2}+\alpha L^u_{S2}}$ \\
				\quad\textbf{Update} ${f_{S1}}$ with ${L_{S1}}$, and \textbf{Update} ${f_{S2}}$ with ${L_{S2}}$ \\
				\quad\textbf{Update} ${f_{T1}}$ and ${f_{T2}}$ via EMA \\
				\textbf{end for}
			\EndFor
		\end{algorithmic}
	\end{flushleft}
\end{algorithm}

\begin{table*}[!h]
  \small
	\begin{tabular}{lc|r@{$\pm$}l|r@{$\pm$}l|r@{$\pm$}l|r@{$\pm$}l|c}
		\toprule
		Methods & Reference & \multicolumn{2}{c|}{1\% COCO} & \multicolumn{2}{c|}{2\% COCO} & \multicolumn{2}{c|}{5\% COCO} & \multicolumn{2}{c|}{10\% COCO} & 100\% COCO \\
    \midrule
    \midrule
    Supervised & - & \multicolumn{2}{c|}{9.05$\pm$0.16} & \multicolumn{2}{c|}{12.70$\pm$0.15} & \multicolumn{2}{c|}{18.47$\pm$0.22} & \multicolumn{2}{c|}{23.86$\pm$0.81} & 37.63 \\
		\midrule
    CSD \cite{jeong2019consistency} & NeurIPS-19 & 10.51 & 0.06 \small{(+1.46)} & 13.93 & 0.12 \small{(+1.23)} & 18.63 & 0.07 \small{(+0.16)} & 22.46 & 0.08 \small{(\ -1.40)} & 38.87 \small{(+1.24)} \\
		STAC \cite{sohn2020simple} & ArXiv-20 & 13.97 & 0.35 \small{(+4.92)} & 18.25 & 0.25 \small{(+5.55)} & 24.38 & 0.12 \small{(+5.86)} & 28.64 & 0.21 \small{(+4.78)} & 39.21 \small{(+1.58)} \\
    Instant-Teaching \cite{zhou2021instant} & CVPR-21 & 18.05 & 0.15 \small{(+9.00)} & 22.45 & 0.15 \small{(+9.75)} & 26.75 & 0.05 \small{(+8.28)} & 30.40 & 0.05 \small{(+6.54)} & 40.20 \small{(+2.57)} \\
    ISMT \cite{yang2021interactive} & CVPR-21 & 18.88 & 0.74 \small{(+9.83)} & 22.43 & 0.56 \small{(+9.73)} & 26.37 & 0.24 \small{(+7.90)} & 30.53 & 0.52 \small{(+6.67)} & 39.64 \small{(+2.01)} \\
    Humble Teacher \cite{tang2021humble} & CVPR-21 & 16.96 & 0.38 \small{(+7.91)} & 21.72 & 0.24 \small{(+9.02)} & 27.70 & 0.75 \small{(+9.23)} & 31.61 & 0.28 \small{(+7.75)} & 42.37 \small{(+4.74)} \\
    Combating Noise \cite{wang2021combating} & NeurIPS-21 & 18.41 & 0.10 \small{(+9.36)} & 24.00 & 0.15 \small{(+11.30)} & 28.96 & 0.29 \small{(+10.49)} & 32.43 & 0.20 \small{(+8.57)} & 43.20 \small{(+5.57)} \\
    Soft Teacher \cite{xu2021end} & ICCV-21 & 20.46 & 0.39 \small{(+11.41)} &  \multicolumn{2}{c|}{-} & \textbf{30.74} & \textbf{0.08} \textbf{\small{(+12.27)}} & \textbf{34.04} & \textbf{0.14} \textbf{\small{(+10.18)}} & \textbf{44.50} \textbf{\small{(+6.87)}} \\
    CPL \cite{li2021rethinking} & AAAI-22 & 19.02 & 0.25 \small{(+9.97)} & 23.34 & 0.18 \small{(+10.64)} & 28.40 & 0.15 \small{(+9.93)} & 32.23 & 0.14 \small{(+8.37)} & 43.30 \small{(+5.67)} \\
    MUM \cite{kim2021mum} & CVPR-22 & \textbf{21.88} & \textbf{0.12} \textbf{\small{(+12.83)}} &\textbf{24.84} & \textbf{0.10} \textbf{\small{(+12.14)}} & 28.52 & 0.09 \small{(+10.05)} & 31.87 & 0.30 \small{(+8.01)} & 42.11 \small{(+4.48)} \\
    \midrule
    Unbiased Teacher \cite{liu2020unbiased} & ICLR21 & 20.75 & 0.12 \small{(+11.70)} & 24.30 & 0.07 \small{(+11.60)} & 28.27 & 0.11 \small{(+9.80)} & 31.50 & 0.10 \small{(+7.64)} & 41.30 \small{(+3.67)} \\
    \textbf{CST\;\, (ours)} & - & 22.20 & 0.18 \small{(+13.15)} & 26.17 & 0.15 \small{(+13.47)} & 29.75 & 0.13 \small{(+11.28)} & 32.65 & 0.21 \small{(+8.79)} & 42.05 \small{(+4.42)} \\
    \textbf{CST* (ours)} & - & \textbf{22.73} & \textbf{0.14} \small{\textbf{(+13.68)}} & \textbf{26.94} & \textbf{0.10} \small{\textbf{(+14.24)}} & \textbf{30.83} & \textbf{0.08} \small{\textbf{(+12.36)}} & \textbf{33.90} & \textbf{0.17} \small{\textbf{(+10.04)}} & \textbf{43.37} \small{\textbf{(+5.74)}}\\
    \bottomrule
	\end{tabular}
  \caption{The performance (AP\%) of different semi-supervised object detection methods for 1\%, 2\%, 5\%, 10\% and 100\% MS-COCO protocols. All methods use ResNet-50 with FPN as backbone and Unbiased Teacher is treated as baseline for a fair comparison.} 
  \label{tab:performance}
\end{table*}

\textbf{Reweighted Loss.} After the process of consistency quantification, the values for all pseudo boxes can be acquired. Then the foreground box candidates generated from S2 are assigned with corresponding quantification values during the process of label assignment. Based on the assigned values, the foreground classification loss can be re-weighted to mitigate the accumulated biases. Given two box sets ${\{x_i^{f}\}}$ and ${\{x_i^{b}\}}$, with ${\{x_i^{f}\}}$ denoting boxes assigned as foreground and ${\{x_i^{b}\}}$ denoting the boxes assigned as background, the classification loss of an unlabeled image with the consistency weighting can be defined as follows:
\begin{equation}
	L_u^{cls} =  \frac{1}{N_f}\sum_{i=1}^{N_f}c_{i}L_{cls}(x_i^f,\hat{y}_i^f)+\frac{1}{N_b}\sum_{i=1}^{N_b}L_{cls}(x_i^b,\hat{y}_i^b)
\end{equation}

where ${N_f}$ and ${N_b}$ are the number of box candidates of the box set ${\{x_i^f\}}$ and ${\{x_i^b\}}$ respectively.
${\hat{y}_i^f}$ and ${\hat{y}_i^b}$ denotes the categories of pseudo boxes generated from teacher T1. ${L_{cls}}$ is the box classification loss. Including the above distribution consistency reweighting strategy, the whole training process is described in Algorithm \ref{alg}.

\section{Experiments}



\subsection{Experimental Settings}

\textbf{Dataset.} MS-COCO \cite{lin2014microsoft} and PASCAL VOC \cite{everingham2010pascal} 
datasets are used in our experiments following the previous SSOD works. 
MS-COCO dataset contains 118k labeled images for training with approximate 850k instances of 80 categories. 
PASCAL VOC 2007 dataset contains 5k labeled images for training with 24k instances of 20 categories, 
while PASCAL VOC 2012 dataset contains 11.5k labeled images for training with 27k instances.
For the experiments on MS-COCO, $0.5\%$, $1\%$, $2\%$, $5\%$ and $10\%$ of the labeled training data are randomly sampled 
and the remainder is taken as the unlabeled data. In addition, we also use \textit{coco-full} dataset for the 100\% protocol, 
which is composed of the 118k standard training set of MS-COCO as labeled dataset and the 120k COCO2017 unlabeled data 
as the unlabeled training set, 
to further measure the effect of our framework. For the experiments on PASCAL VOC, VOC 2007 dataset 
is utilized as the labeled training set and VOC 2012 dataset as the unlabeled training set. 
Moreover, we add the images from COCO dataset that share the same 20 object categories with VOC 2007 dataset to the unlabeled training set as \textit{VOC-additional} dataset.
The detection performance is evaluated on COCO2017-val set for the COCO dataset 
and VOC 2007-test set for the VOC dataset following the existing works. 

\begin{table}[!t]
  \small
	\begin{tabular}{lc|r@{$\pm$}l}
		\toprule
		Methods & Reference & \multicolumn{2}{c}{0.5\% COCO} \\
    \midrule
    \midrule
    Supervised & - & \multicolumn{2}{c}{6.83$\pm$0.15} \\
    \midrule
    CSD \cite{jeong2019consistency} & NeurIPS-19 & 7.41 & 0.21 \small{(+0.58)} \\
    STAC \cite{sohn2020simple} & ArXiv-20 & 9.78 & 0.53 \small{(+2.95)} \\
    MUM \cite{kim2021mum} & CVPR-22 & \textbf{18.54} & \textbf{0.48} \textbf{\small{(+11.71)}} \\
    \midrule
    Unbiased Teacher \cite{liu2020unbiased} & ICLR-21 & 16.94 & 0.23 \small{(+10.11)} \\
    \textbf{CST\;\, (ours)} & - & 19.20 & 0.28 \small{(+12.37)} \\
    \textbf{CST* (ours)} & - & \textbf{19.65} & \textbf{0.21 \small{(+12.82)}}  \\
    \bottomrule
	\end{tabular}
  \caption{The performance (AP\%) of existing SSOD methods with ResNet-50-FPN backbone for 0.5\% MS-COCO protocol.}
  \label{tab:small_res}
\end{table}
\noindent\textbf{Implementation Details.} Faster R-CNN \cite{FasterRCNN} with  
ResNet-50-FPN \cite{he2016deep, lin2017feature} backbone is involved in our experiments to
ensure the fairness and correctness in comparisons followed by the existing works \cite{sohn2020simple,liu2020unbiased,yang2021interactive,xu2021end,wang2021combating,kim2021mum}. 
All hyper-parameters and augmentations are reserved as Unbiased Teacher \cite{liu2020unbiased} and 
the weights of the model are initialized from the pre-trained models on ImageNet \cite{russakovsky2015imagenet}. 
We use an SGD optimizer on 8 GPUs with a learning rate 0.01, a momentum rate 0.9 and a weight decay 0.0001. 
The batch size is set to 32 in our main experiments compared to other existing methods.
For 0.5\%, 1\%, 2\%, 5\% and 10\% MS-COCO protocols, we adopt 180k training iterations, including 
1k, 2k, 6k, 12k and 20k iterations for the initial Burn-in stage that only train the model by labeled data 
and the rest for the cycle self-training stage. 
Specially, for 100\% (\textit{COCO-full}) protocol, the Burn-in stage is set to 90k and the total number of 
training iterations is 360k. 
For PASCAL VOC dataset and \textit{VOC-additional} dataset, the Burn-in stage is set to 30k and the whole training iterations 
are 180k. 
Following the convention, the 
confidence score threshold is set to 0.7 for filtering the pseudo labels. For the training loss, we set the weight factor $\alpha=4$ for the unlabeled data following the Unbiased Teacher \cite{liu2020unbiased}. 
Specially, during the testing process, apart from the detection results produced by the teacher model, 
we also evaluate the ensemble results of the two teacher models via Weighted Boxes Fusion (WBF) \cite{solovyev2021weighted}, which further improves the performance of our proposed method.  

\subsection{Comparisons with State-of-the-art}

\begin{figure*}[t]
	\centering
	\includegraphics[width=0.85\textwidth]{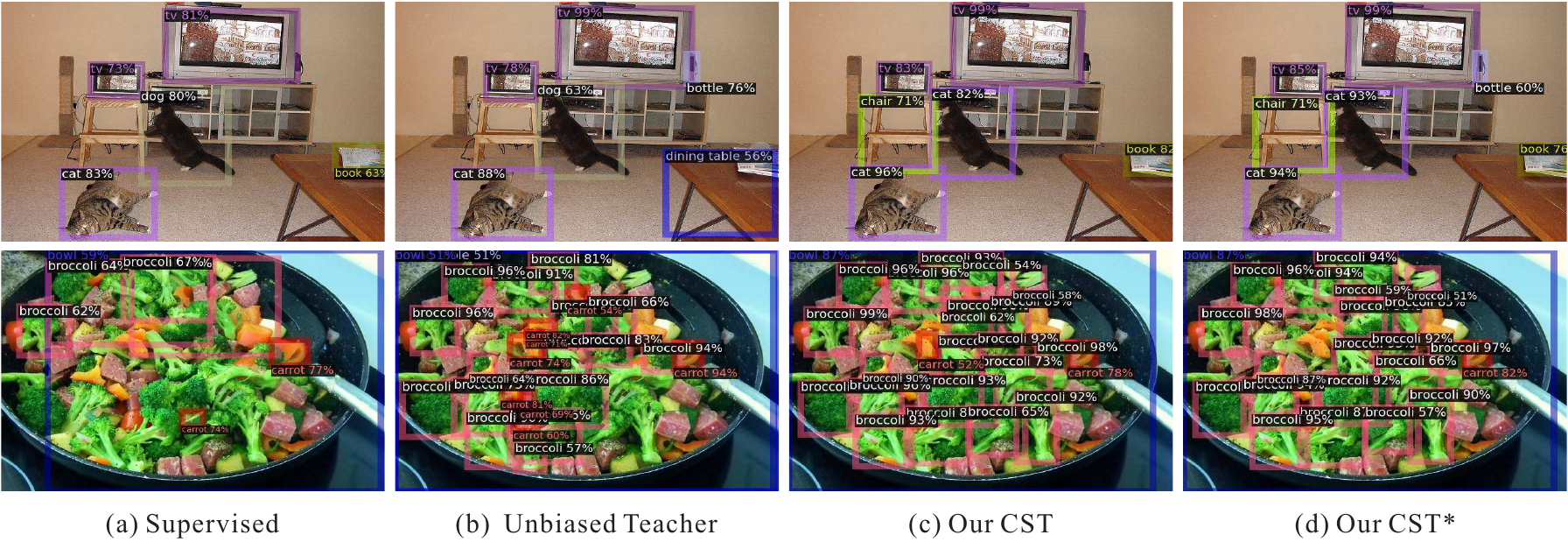}
	\caption{Visualization results of different methods. (a) Supervised. (b) Unbiased Teacher. (c) Our CST. (d) Our CST*.}
  \label{fig:result}
\end{figure*}

\textbf{MS-COCO.} We first preform the comparative study on MS-COCO dataset to evaluate our CST and CST* method. 
The results are reported in Table \ref{tab:performance}. For the 1\%, 2\% and 5\% protocol, our method outperforms all the state-of-the-art methods. 
Compared to the baseline Unbiased Teacher \cite{liu2020unbiased}, our CST* achieves 22.73\% AP, 26.94\% and 30.83\% AP with only 1\%, 2\% and 5\% unlabeled data respectively. 
For 10\% and 100\% MS-COCO protocols, our CST* improves 2.40\%, 2.07\% AP over the baseline Unbiased Teacher, which also 
surpasses most of the methods, including the recent works CPL \cite{li2021rethinking} and MUM \cite{kim2021mum}. 
These results show that our proposed CST and CST* is reliable and outstanding in semi-supervised object detection. 

In order to illustrate that our CST and CST* can give higher AP results when less labeled data are provided, 
we conduct the experiment on 0.5\% MS-COCO dataset. The evaluation results 
are shown in Table \ref{tab:small_res}, CST* achieves 19.65\% AP 
and outperforms the best of existing work. 
The reason is that with less labeled samples, due to the worse pre-trained model in the burn-in stage, 
the pseudo-labels generated from the teacher become more and more unstable and inaccurate. 
The gap of the predictions is increased between the two teachers, so the proportion of reweighted 
foreground bounding box candidates increases, which facilitates the performance gains caused by distribution consistency reweighting. 

\begin{table}[t]
  \small
	\begin{tabular}{lc|cc|cc}
		\toprule
		\multirow{2}{*}{Methods} & \multirow{2}{*}{Reference} & \multicolumn{2}{c|}{PASCAL VOC} & \multicolumn{2}{c}{VOC-additional} \\ 
    & & $AP$ & $AP_{50}$ & $AP$ & $AP_{50}$ \\
    \midrule
    \midrule
    Supervised & - & 45.3 & 76.3 & 45.3 & 76.3 \\
    CSD \cite{jeong2019consistency} & NeurIPS-19 & - & 74.7 & - & 75.1 \\
    STAC \cite{sohn2020simple} & ArXiv-20 & 44.6 & 77.4 & 46.0 & 79.1 \\
    Instant-Teaching \cite{zhou2021instant} & CVPR-21 & 48.7 & 78.3 & 49.7 & 79.0 \\
    ISMT \cite{yang2021interactive} & CVPR-21 & 46.2 & 77.2 & 49.6 & 77.7 \\
    Humble Teacher \cite{tang2021humble} & CVPR-21 & \textbf{53.0} & \textbf{80.9} & \textbf{54.4} & 81.3 \\
    Combating Noise \cite{wang2021combating} & NeurIPS-21 & 49.3 & 80.6 & 50.2 & \textbf{81.4} \\
    CPL \cite{li2021rethinking} & AAAI-22 & 52.4 & 76.9 & 54.0 & 77.6 \\
    MUM \cite{kim2021mum} & CVPR-22 & 50.2 & 78.9 & 52.3 & 80.5 \\
    \midrule
    Unbiased Teacher \cite{liu2020unbiased} & ICLR-21 & 48.7 & 77.4 & 50.3 & 78.8 \\
    \textbf{CST\;\, (ours)} & - & 50.3 & 78.1 & 52.3 & 79.6 \\
    \textbf{CST* (ours)} & - & \textbf{51.5} & \textbf{78.7} & \textbf{53.5} & \textbf{80.5} \\
    \bottomrule
	\end{tabular}
  \caption{The performance of existing SSOD methods for PASCAL VOC dataset and VOC-additional dataset.}
  \label{tab:voc_perform}
\end{table}

\noindent\textbf{PASCAL VOC.} Then we evaluate our proposed CST and CST* on PASCAL VOC and \textit{VOC-additional} dataset. 
As shown in Table \ref{tab:voc_perform}, our method still preforms better compared to the baseline. For the first protocol, 
CST* achieves about 51.5\% for AP and 78.7\% for AP$_{50}$ improvement over the baseline. 
For the second protocol, 
CST* improves from 50.3\% to 53.5\% for AP and from 78.8\% to 80.5\% for AP$_{50}$. These evaluation results 
indicate that our CST* still achieves the comparable performance on the semi-supervised object detection 
datasets with less categories. In addition, we find that our CST* preforms better on MS-COCO than PASCAL VOC. 
This is because PASCAL VOC is so simple due to the less images and categories that probably cause the over-fitting problem, 
Consequently, this observation confirms our conclusion that unlabeled data with more knowledge and patterns can improve the effect of our method.


\subsection{Qualitative Results}

We give some visual results for comparison, including the supervised method, Unbiased Teacher, our CST and CST*. 
As shown in Figure \ref{fig:result}, these detection results of objects from different categories 
obviously manifest the superiority of our proposed method. For instance, 
the chair can be better detected and and the cat can be correctly categorized by our CST
Additionally, the figure also shows that the effect of the detection results can be further improved by our CST* method. For example, most of broccoli can be detected by our CST and CST* while less are presented by the Unbiased Teacher.
Based on these visualization results, our CST and CST* can achieve more precise classification and accurate localization results. 

\begin{table*}[htbp]
  \centering
  \begin{minipage}{0.3\linewidth}
    \centering
    \begin{tabular}{lc|c}
      \toprule
      CST & DCR & AP (\%) \\
      \midrule
      &  & 20.1 \\
      \checkmark &  & 21.4 \\
      & \checkmark & 21.2 \\
      \checkmark & \checkmark & \textbf{21.9} \\
      \bottomrule
    \end{tabular}
    \caption{Ablation study of different network components.}
    \label{tab:component}
  \end{minipage}
  \quad
  \begin{minipage}{0.3\linewidth}
    \centering
    \begin{tabular}{c|c}
      \toprule
      Quantification Style & AP (\%) \\
      \midrule
      JS Divergence & 21.5 \\
      L1 Distance & \textbf{21.9} \\
      \bottomrule
    \end{tabular}
    \caption{Ablation study of consistency quantification style in DCR.}
    \label{tab:style}
  \end{minipage}
  \begin{minipage}{0.33\linewidth}
    \centering
    \begin{tabular}{c|c}
      \toprule
      Combinations for DCR & AP (\%) \\
      \midrule
      (T1 $\leftrightarrow$ S1) ,\quad (T2 $\leftrightarrow$ S2) & 21.6 \\
      (T1 $\leftrightarrow$ S2) ,\quad (T2 $\leftrightarrow$ S1) & 21.4 \\
      (T1 $\leftrightarrow$ T2) ,\quad (T2 $\leftrightarrow$ T1) & \textbf{21.9} \\
      \bottomrule
    \end{tabular}
    \caption{Ablation study of different combinations in DCR.} 
    \label{tab:combination}
  \end{minipage}
\end{table*}

\begin{figure*}
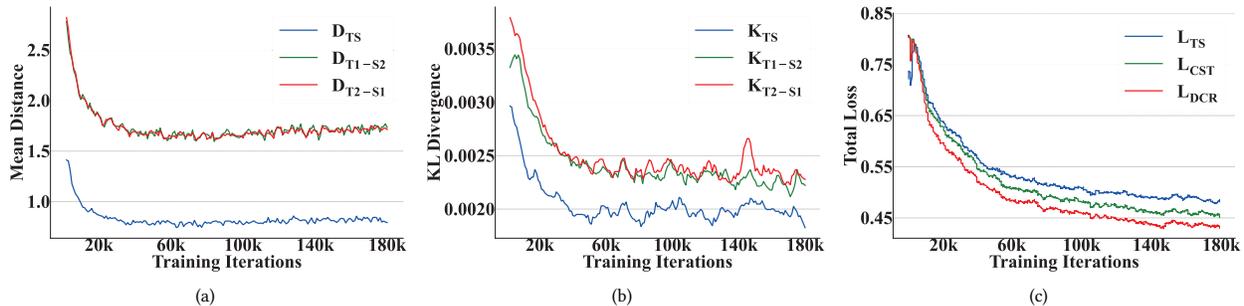

	\vspace{-15pt}
	\centering
	\subfigure[]{
		\begin{minipage}[b]{0.298\textwidth}
				\includegraphics[width=\textwidth]{Figure/roi}
		\end{minipage}
	}
	\subfigure[]{
		\begin{minipage}[b]{0.298\textwidth}
				\includegraphics[width=\textwidth]{Figure/kl}
		\end{minipage}
	}
	\subfigure[]{
		\begin{minipage}[b]{0.298\textwidth}
				\includegraphics[width=\textwidth]{Figure/loss}
		\end{minipage}
	}
	\vspace{-10pt}
	\caption{(a) Mean Distance between the features of the same region for different components configurations. (b) KL divergence between the classification predictions of the same region. (c) Training Loss curves of different components.}
  \label{fig:discuss}
\end{figure*}

\subsection{Ablation Studies}

We conduct our ablation studies on 1\% MS-COCO dataset to evaluate the effectiveness of each component in our CST framework. Without loss of fairness, we use 16 batch size in our ablation studies because of lower computation and faster training process. 

\textbf{Effects of Each Component.}
Our main contribution contains two components: the Cycle Self-Training (CST) framework and the Distribution Consistency Reweighting (DCR) strategy. 
To validate the effect of the two components, we take the Unbiased Teacher as the baseline, and then evaluate various component combinations as shown in Table \ref{tab:component}. 
We can observe that adding either of the two components can give a favorable improvement. 
For the CST component AP gains 1.3\% and for the DCR component AP improves 1.1\%. 
By incorporating both of the two components, the performance is significantly boosted to 21.9\% compared to the baseline 20.1\%. 

\textbf{Different Consistency Quantification Styles.}
Since there are mainly two styles of distribution consistency quantification as mentioned earlier, 
the experiments with different consistency quantification styles are conducted to verify which style is more compatible with our proposed framework. 
Table \ref{tab:style} shows the detailed evaluation results while using different styles during training. 
From the table we can conclude that the L1 distance between the two classification predictions performs better than the JS divergence when treated as consistency quantification style in our framework. 
The reason is quite likely to be that the L1 distance 
better balances the weights of different positive samples 
compared to the JS divergence. 

\textbf{Different Consistency.}
Although we have analyzed the effectiveness of the DCR component, all previous experiments compute the consistency between the two teacher models. 
Beyond the (T1 $\leftrightarrow$ T2), (T2 $\leftrightarrow$ T1) setting, i.e., the consistency computation between the teacher T1 and the teacher T2, we wonder that 
what if we compute the consistency between one teacher model and another student model. Thus we conduct the ablation studies on different settings 
of combinations between (T1 $\leftrightarrow$ S1), (T2 $\leftrightarrow$ S2) and (T1 $\leftrightarrow$ S2), (T2 $\leftrightarrow$ S1). 
The final results are given in Table \ref{tab:combination}. Both the first two combinations give good performance, 
however, as what we expected, the computation between T1 and T2 achieves 21.9\% AP and presents the best performance among them, 
which supports our practice of calculating the distribution consistency between the two teacher models to mitigate the confirmation biases problem. 


\subsection{Discussions}

\textbf{Coupling Effect.} To verify our proposed CST framework indeed helps to overcome the tightly coupling 
effect, we visualize the Euclidean distances between the features of the same regions extracted by 
the teacher T1 (T2) and the student S2 (S1). Specifically, given the pseudo boxes generated from T1 (T2), 
we can obtain the aligned features from S2 (S1) through the detection head. Then the mean of Euclidean 
distances between the features extracted by T1 and S2 is calculated, which is denoted as ${D_{T1-S2}}$. 
Similarly, the mean distance ${D_{T2-S1}}$ and the mean distance ${D_{T-S}}$ of the teacher-student 
framework are also provided, as shown in Figure \ref{fig:discuss} (a). 
Compared with ${D_{T-S}}$, the mean distance ${D_{T1-S2}}$ and ${D_{T2-S1}}$ are always larger during 
the training process, which proves that T1(T2) and S2(S1) are not tightly coupled similar to T and S. 
By doing so, the performance limitation caused by the conventional teacher-student framework can be 
broken. 
At the same time, the KL divergence of the classification predictions of the teacher T1 (T2) and the student S2 (S1) 
is visualized in Figure \ref{fig:discuss} (b), which also indicates the coupled problem 
can be alleviated through our CST framework.



\noindent\textbf{Distribution Consistency Reweighting.} To verify the effectiveness of our proposed distribution consistency reweighting strategy, we present training curves of different component configurations in Figure \ref{fig:discuss} (c), including the conventional teacher-student framework, the proposed CST framework and the CST framework with distribution consistency reweighting strategy, denoted as ${L_{TS}}$, ${L_{CST}}$ and ${L_{DCR}}$ respectively. We observe that, with our proposed CST framework, the model has lower losses compared with the original teacher-student framework. When we utilize the overall framework, it achieves the lowest losses, which indicates that the model can be trained more robustly with noisy pseudo labels and mitigate accumulating confirmation biases under our proposed method.

\section{Conclusion}

In this paper, we propose a Cycle Self-Training (CST) framework for semi-supervised object detection, in which a knowledge transferring loop is built to loose the tightly coupling effect of the conventional teacher-student framework. 
Furthermore, a distribution consistency reweighting (DCR) strategy is introduced to be combined with the proposed CST framework to train the student 
networks robustly with noisy pseudo labels to avoid accumulating confirmation biases. 
Extensive experiments on MS-COCO and PASCAL VOC datasets demonstrate the effectiveness of our proposed 
framework. Moreover, the proposed framework is fairly general and can be easily incorporated with 
existing object detection methods to perform the semi-supervised learning.

\begin{acks}
This work is supported in part by State Grid Corporation of China Headquarters Project which is "Research on small sample training method based on deep neural network and its application in power system" under Grant 5400-202158333A-0-0-00. Also it is supported by National Key R\&D Program of China (2018YFB0804203) and National Natural Science Foundation of China (62072438,U1936110).
\end{acks}

\bibliographystyle{ACM-Reference-Format}
\bibliography{sample-base}


\begin{thebibliography}{49}


\ifx \showCODEN    \undefined \def \showCODEN     #1{\unskip}     \fi
\ifx \showDOI      \undefined \def \showDOI       #1{#1}\fi
\ifx \showISBNx    \undefined \def \showISBNx     #1{\unskip}     \fi
\ifx \showISBNxiii \undefined \def \showISBNxiii  #1{\unskip}     \fi
\ifx \showISSN     \undefined \def \showISSN      #1{\unskip}     \fi
\ifx \showLCCN     \undefined \def \showLCCN      #1{\unskip}     \fi
\ifx \shownote     \undefined \def \shownote      #1{#1}          \fi
\ifx \showarticletitle \undefined \def \showarticletitle #1{#1}   \fi
\ifx \showURL      \undefined \def \showURL       {\relax}        \fi
\providecommand\bibfield[2]{#2}
\providecommand\bibinfo[2]{#2}
\providecommand\natexlab[1]{#1}
\providecommand\showeprint[2][]{arXiv:#2}

\bibitem[Assran et~al\mbox{.}(2021)]%
        {assran2021semi}
\bibfield{author}{\bibinfo{person}{Mahmoud Assran}, \bibinfo{person}{Mathilde
  Caron}, \bibinfo{person}{Ishan Misra}, \bibinfo{person}{Piotr Bojanowski},
  \bibinfo{person}{Armand Joulin}, \bibinfo{person}{Nicolas Ballas}, {and}
  \bibinfo{person}{Michael Rabbat}.} \bibinfo{year}{2021}\natexlab{}.
\newblock \showarticletitle{Semi-supervised learning of visual features by
  non-parametrically predicting view assignments with support samples}. In
  \bibinfo{booktitle}{\emph{Proceedings of the IEEE/CVF International
  Conference on Computer Vision}}. \bibinfo{pages}{8443--8452}.
\newblock


\bibitem[Berthelot et~al\mbox{.}(2019)]%
        {berthelot2019remixmatch}
\bibfield{author}{\bibinfo{person}{David Berthelot}, \bibinfo{person}{Nicholas
  Carlini}, \bibinfo{person}{Ekin~D Cubuk}, \bibinfo{person}{Alex Kurakin},
  \bibinfo{person}{Kihyuk Sohn}, \bibinfo{person}{Han Zhang}, {and}
  \bibinfo{person}{Colin Raffel}.} \bibinfo{year}{2019}\natexlab{}.
\newblock \showarticletitle{ReMixMatch: Semi-Supervised Learning with
  Distribution Matching and Augmentation Anchoring}. In
  \bibinfo{booktitle}{\emph{International Conference on Learning
  Representations}}.
\newblock


\bibitem[Bochkovskiy et~al\mbox{.}(2020)]%
        {bochkovskiy2020yolov4}
\bibfield{author}{\bibinfo{person}{Alexey Bochkovskiy},
  \bibinfo{person}{Chien-Yao Wang}, {and} \bibinfo{person}{Hong-Yuan~Mark
  Liao}.} \bibinfo{year}{2020}\natexlab{}.
\newblock \showarticletitle{Yolov4: Optimal speed and accuracy of object
  detection}.
\newblock \bibinfo{journal}{\emph{arXiv preprint arXiv:2004.10934}}
  (\bibinfo{year}{2020}).
\newblock


\bibitem[Cai and Vasconcelos(2018)]%
        {cai2018cascade}
\bibfield{author}{\bibinfo{person}{Zhaowei Cai} {and} \bibinfo{person}{Nuno
  Vasconcelos}.} \bibinfo{year}{2018}\natexlab{}.
\newblock \showarticletitle{Cascade r-cnn: Delving into high quality object
  detection}. In \bibinfo{booktitle}{\emph{Proceedings of the IEEE conference
  on computer vision and pattern recognition}}. \bibinfo{pages}{6154--6162}.
\newblock


\bibitem[Carion et~al\mbox{.}(2020)]%
        {carion2020end}
\bibfield{author}{\bibinfo{person}{Nicolas Carion}, \bibinfo{person}{Francisco
  Massa}, \bibinfo{person}{Gabriel Synnaeve}, \bibinfo{person}{Nicolas
  Usunier}, \bibinfo{person}{Alexander Kirillov}, {and} \bibinfo{person}{Sergey
  Zagoruyko}.} \bibinfo{year}{2020}\natexlab{}.
\newblock \showarticletitle{End-to-end object detection with transformers}. In
  \bibinfo{booktitle}{\emph{European conference on computer vision}}. Springer,
  \bibinfo{pages}{213--229}.
\newblock


\bibitem[Chen et~al\mbox{.}(2019)]%
        {chen2019hybrid}
\bibfield{author}{\bibinfo{person}{Kai Chen}, \bibinfo{person}{Jiangmiao Pang},
  \bibinfo{person}{Jiaqi Wang}, \bibinfo{person}{Yu Xiong},
  \bibinfo{person}{Xiaoxiao Li}, \bibinfo{person}{Shuyang Sun},
  \bibinfo{person}{Wansen Feng}, \bibinfo{person}{Ziwei Liu},
  \bibinfo{person}{Jianping Shi}, \bibinfo{person}{Wanli Ouyang},
  {et~al\mbox{.}}} \bibinfo{year}{2019}\natexlab{}.
\newblock \showarticletitle{Hybrid task cascade for instance segmentation}. In
  \bibinfo{booktitle}{\emph{Proceedings of the IEEE/CVF Conference on Computer
  Vision and Pattern Recognition}}. \bibinfo{pages}{4974--4983}.
\newblock


\bibitem[Dai et~al\mbox{.}(2016)]%
        {dai2016r}
\bibfield{author}{\bibinfo{person}{Jifeng Dai}, \bibinfo{person}{Yi Li},
  \bibinfo{person}{Kaiming He}, {and} \bibinfo{person}{Jian Sun}.}
  \bibinfo{year}{2016}\natexlab{}.
\newblock \showarticletitle{R-fcn: Object detection via region-based fully
  convolutional networks}.
\newblock \bibinfo{journal}{\emph{Advances in neural information processing
  systems}}  \bibinfo{volume}{29} (\bibinfo{year}{2016}).
\newblock


\bibitem[Dong et~al\mbox{.}(2020)]%
        {dong2020centripetalnet}
\bibfield{author}{\bibinfo{person}{Zhiwei Dong}, \bibinfo{person}{Guoxuan Li},
  \bibinfo{person}{Yue Liao}, \bibinfo{person}{Fei Wang},
  \bibinfo{person}{Pengju Ren}, {and} \bibinfo{person}{Chen Qian}.}
  \bibinfo{year}{2020}\natexlab{}.
\newblock \showarticletitle{Centripetalnet: Pursuing high-quality keypoint
  pairs for object detection}. In \bibinfo{booktitle}{\emph{Proceedings of the
  IEEE/CVF conference on computer vision and pattern recognition}}.
  \bibinfo{pages}{10519--10528}.
\newblock


\bibitem[Duan et~al\mbox{.}(2019)]%
        {duan2019centernet}
\bibfield{author}{\bibinfo{person}{Kaiwen Duan}, \bibinfo{person}{Song Bai},
  \bibinfo{person}{Lingxi Xie}, \bibinfo{person}{Honggang Qi},
  \bibinfo{person}{Qingming Huang}, {and} \bibinfo{person}{Qi Tian}.}
  \bibinfo{year}{2019}\natexlab{}.
\newblock \showarticletitle{Centernet: Keypoint triplets for object detection}.
  In \bibinfo{booktitle}{\emph{Proceedings of the IEEE/CVF international
  conference on computer vision}}. \bibinfo{pages}{6569--6578}.
\newblock


\bibitem[Everingham et~al\mbox{.}(2010)]%
        {everingham2010pascal}
\bibfield{author}{\bibinfo{person}{Mark Everingham}, \bibinfo{person}{Luc
  Van~Gool}, \bibinfo{person}{Christopher~KI Williams}, \bibinfo{person}{John
  Winn}, {and} \bibinfo{person}{Andrew Zisserman}.}
  \bibinfo{year}{2010}\natexlab{}.
\newblock \showarticletitle{The pascal visual object classes (voc) challenge}.
\newblock \bibinfo{journal}{\emph{International journal of computer vision}}
  \bibinfo{volume}{88}, \bibinfo{number}{2} (\bibinfo{year}{2010}),
  \bibinfo{pages}{303--338}.
\newblock


\bibitem[He et~al\mbox{.}(2016)]%
        {he2016deep}
\bibfield{author}{\bibinfo{person}{Kaiming He}, \bibinfo{person}{Xiangyu
  Zhang}, \bibinfo{person}{Shaoqing Ren}, {and} \bibinfo{person}{Jian Sun}.}
  \bibinfo{year}{2016}\natexlab{}.
\newblock \showarticletitle{Deep residual learning for image recognition}. In
  \bibinfo{booktitle}{\emph{Proceedings of the IEEE conference on computer
  vision and pattern recognition}}. \bibinfo{pages}{770--778}.
\newblock


\bibitem[Hu et~al\mbox{.}(2021)]%
        {hu2021simple}
\bibfield{author}{\bibinfo{person}{Zijian Hu}, \bibinfo{person}{Zhengyu Yang},
  \bibinfo{person}{Xuefeng Hu}, {and} \bibinfo{person}{Ram Nevatia}.}
  \bibinfo{year}{2021}\natexlab{}.
\newblock \showarticletitle{Simple: Similar pseudo label exploitation for
  semi-supervised classification}. In \bibinfo{booktitle}{\emph{Proceedings of
  the IEEE/CVF Conference on Computer Vision and Pattern Recognition}}.
  \bibinfo{pages}{15099--15108}.
\newblock


\bibitem[Jeong et~al\mbox{.}(2019)]%
        {jeong2019consistency}
\bibfield{author}{\bibinfo{person}{Jisoo Jeong}, \bibinfo{person}{Seungeui
  Lee}, \bibinfo{person}{Jeesoo Kim}, {and} \bibinfo{person}{Nojun Kwak}.}
  \bibinfo{year}{2019}\natexlab{}.
\newblock \showarticletitle{Consistency-based semi-supervised learning for
  object detection}.
\newblock \bibinfo{journal}{\emph{Advances in neural information processing
  systems}}  \bibinfo{volume}{32} (\bibinfo{year}{2019}).
\newblock


\bibitem[Ke et~al\mbox{.}(2019)]%
        {ke2019dual}
\bibfield{author}{\bibinfo{person}{Zhanghan Ke}, \bibinfo{person}{Daoye Wang},
  \bibinfo{person}{Qiong Yan}, \bibinfo{person}{Jimmy Ren}, {and}
  \bibinfo{person}{Rynson~WH Lau}.} \bibinfo{year}{2019}\natexlab{}.
\newblock \showarticletitle{Dual student: Breaking the limits of the teacher in
  semi-supervised learning}. In \bibinfo{booktitle}{\emph{Proceedings of the
  IEEE/CVF International Conference on Computer Vision}}.
  \bibinfo{pages}{6728--6736}.
\newblock


\bibitem[Kim et~al\mbox{.}(2021)]%
        {kim2021mum}
\bibfield{author}{\bibinfo{person}{JongMok Kim}, \bibinfo{person}{Jooyoung
  Jang}, \bibinfo{person}{Seunghyeon Seo}, \bibinfo{person}{Jisoo Jeong},
  \bibinfo{person}{Jongkeun Na}, {and} \bibinfo{person}{Nojun Kwak}.}
  \bibinfo{year}{2021}\natexlab{}.
\newblock \showarticletitle{MUM: Mix Image Tiles and UnMix Feature Tiles for
  Semi-Supervised Object Detection}.
\newblock \bibinfo{journal}{\emph{arXiv preprint arXiv:2111.10958}}
  (\bibinfo{year}{2021}).
\newblock


\bibitem[Kim et~al\mbox{.}(2022)]%
        {kim2022semi}
\bibfield{author}{\bibinfo{person}{Jiwon Kim}, \bibinfo{person}{Kwangrok Ryoo},
  \bibinfo{person}{Junyoung Seo}, \bibinfo{person}{Gyuseong Lee},
  \bibinfo{person}{Daehwan Kim}, \bibinfo{person}{Hansang Cho}, {and}
  \bibinfo{person}{Seungryong Kim}.} \bibinfo{year}{2022}\natexlab{}.
\newblock \showarticletitle{Semi-Supervised Learning of Semantic Correspondence
  with Pseudo-Labels}. In \bibinfo{booktitle}{\emph{Proceedings of the IEEE/CVF
  Conference on Computer Vision and Pattern Recognition}}.
  \bibinfo{pages}{19699--19709}.
\newblock


\bibitem[Kuznetsova et~al\mbox{.}(2020)]%
        {openimage}
\bibfield{author}{\bibinfo{person}{Alina Kuznetsova}, \bibinfo{person}{Hassan
  Rom}, \bibinfo{person}{Neil Alldrin}, \bibinfo{person}{Jasper R.~R.
  Uijlings}, \bibinfo{person}{Ivan Krasin}, \bibinfo{person}{Jordi
  Pont{-}Tuset}, \bibinfo{person}{Shahab Kamali}, \bibinfo{person}{Stefan
  Popov}, \bibinfo{person}{Matteo Malloci}, \bibinfo{person}{Alexander
  Kolesnikov}, \bibinfo{person}{Tom Duerig}, {and} \bibinfo{person}{Vittorio
  Ferrari}.} \bibinfo{year}{2020}\natexlab{}.
\newblock \showarticletitle{The Open Images Dataset {V4}}.
\newblock \bibinfo{journal}{\emph{Int. J. Comput. Vis.}} \bibinfo{volume}{128},
  \bibinfo{number}{7} (\bibinfo{year}{2020}), \bibinfo{pages}{1956--1981}.
\newblock


\bibitem[Law and Deng(2018)]%
        {law2018cornernet}
\bibfield{author}{\bibinfo{person}{Hei Law} {and} \bibinfo{person}{Jia Deng}.}
  \bibinfo{year}{2018}\natexlab{}.
\newblock \showarticletitle{Cornernet: Detecting objects as paired keypoints}.
  In \bibinfo{booktitle}{\emph{Proceedings of the European conference on
  computer vision (ECCV)}}. \bibinfo{pages}{734--750}.
\newblock


\bibitem[Li et~al\mbox{.}(2021)]%
        {li2021rethinking}
\bibfield{author}{\bibinfo{person}{Hengduo Li}, \bibinfo{person}{Zuxuan Wu},
  \bibinfo{person}{Abhinav Shrivastava}, {and} \bibinfo{person}{Larry~S
  Davis}.} \bibinfo{year}{2021}\natexlab{}.
\newblock \showarticletitle{Rethinking Pseudo Labels for Semi-Supervised Object
  Detection}.
\newblock \bibinfo{journal}{\emph{arXiv preprint arXiv:2106.00168}}
  (\bibinfo{year}{2021}).
\newblock


\bibitem[Lim et~al\mbox{.}(2021)]%
        {lim2021class}
\bibfield{author}{\bibinfo{person}{Jongin Lim}, \bibinfo{person}{Daeho Um},
  \bibinfo{person}{Hyung~Jin Chang}, \bibinfo{person}{Dae~Ung Jo}, {and}
  \bibinfo{person}{Jin~Young Choi}.} \bibinfo{year}{2021}\natexlab{}.
\newblock \showarticletitle{Class-attentive diffusion network for
  semi-supervised classification}. In \bibinfo{booktitle}{\emph{Thirty-Fifth
  AAAI Conference on Artificial Intelligence, AAAI}}. \bibinfo{pages}{2--9}.
\newblock


\bibitem[Lin et~al\mbox{.}(2017a)]%
        {lin2017feature}
\bibfield{author}{\bibinfo{person}{Tsung-Yi Lin}, \bibinfo{person}{Piotr
  Doll{\'a}r}, \bibinfo{person}{Ross Girshick}, \bibinfo{person}{Kaiming He},
  \bibinfo{person}{Bharath Hariharan}, {and} \bibinfo{person}{Serge Belongie}.}
  \bibinfo{year}{2017}\natexlab{a}.
\newblock \showarticletitle{Feature pyramid networks for object detection}. In
  \bibinfo{booktitle}{\emph{Proceedings of the IEEE conference on computer
  vision and pattern recognition}}. \bibinfo{pages}{2117--2125}.
\newblock


\bibitem[Lin et~al\mbox{.}(2017b)]%
        {lin2017focal}
\bibfield{author}{\bibinfo{person}{Tsung-Yi Lin}, \bibinfo{person}{Priya
  Goyal}, \bibinfo{person}{Ross Girshick}, \bibinfo{person}{Kaiming He}, {and}
  \bibinfo{person}{Piotr Doll{\'a}r}.} \bibinfo{year}{2017}\natexlab{b}.
\newblock \showarticletitle{Focal loss for dense object detection}. In
  \bibinfo{booktitle}{\emph{Proceedings of the IEEE international conference on
  computer vision}}. \bibinfo{pages}{2980--2988}.
\newblock


\bibitem[Lin et~al\mbox{.}(2014)]%
        {lin2014microsoft}
\bibfield{author}{\bibinfo{person}{Tsung-Yi Lin}, \bibinfo{person}{Michael
  Maire}, \bibinfo{person}{Serge Belongie}, \bibinfo{person}{James Hays},
  \bibinfo{person}{Pietro Perona}, \bibinfo{person}{Deva Ramanan},
  \bibinfo{person}{Piotr Doll{\'a}r}, {and} \bibinfo{person}{C~Lawrence
  Zitnick}.} \bibinfo{year}{2014}\natexlab{}.
\newblock \showarticletitle{Microsoft coco: Common objects in context}. In
  \bibinfo{booktitle}{\emph{European conference on computer vision}}. Springer,
  \bibinfo{pages}{740--755}.
\newblock


\bibitem[Liu et~al\mbox{.}(2016)]%
        {SSD}
\bibfield{author}{\bibinfo{person}{Wei Liu}, \bibinfo{person}{Dragomir
  Anguelov}, \bibinfo{person}{Dumitru Erhan}, \bibinfo{person}{Christian
  Szegedy}, \bibinfo{person}{Scott Reed}, \bibinfo{person}{Cheng-Yang Fu},
  {and} \bibinfo{person}{Alexander~C. Berg}.} \bibinfo{year}{2016}\natexlab{}.
\newblock \showarticletitle{SSD: Single Shot MultiBox Detector}. In
  \bibinfo{booktitle}{\emph{European Conference on Computer Vision}}.
  \bibinfo{pages}{21--37}.
\newblock


\bibitem[Liu et~al\mbox{.}(2021b)]%
        {liu2020unbiased}
\bibfield{author}{\bibinfo{person}{Yen-Cheng Liu}, \bibinfo{person}{Chih-Yao
  Ma}, \bibinfo{person}{Zijian He}, \bibinfo{person}{Chia-Wen Kuo},
  \bibinfo{person}{Kan Chen}, \bibinfo{person}{Peizhao Zhang},
  \bibinfo{person}{Bichen Wu}, \bibinfo{person}{Zsolt Kira}, {and}
  \bibinfo{person}{Peter Vajda}.} \bibinfo{year}{2021}\natexlab{b}.
\newblock \showarticletitle{Unbiased Teacher for Semi-Supervised Object
  Detection}. In \bibinfo{booktitle}{\emph{International Conference on Learning
  Representations}}.
\newblock


\bibitem[Liu et~al\mbox{.}(2021a)]%
        {liu2021swin}
\bibfield{author}{\bibinfo{person}{Ze Liu}, \bibinfo{person}{Yutong Lin},
  \bibinfo{person}{Yue Cao}, \bibinfo{person}{Han Hu}, \bibinfo{person}{Yixuan
  Wei}, \bibinfo{person}{Zheng Zhang}, \bibinfo{person}{Stephen Lin}, {and}
  \bibinfo{person}{Baining Guo}.} \bibinfo{year}{2021}\natexlab{a}.
\newblock \showarticletitle{Swin transformer: Hierarchical vision transformer
  using shifted windows}. In \bibinfo{booktitle}{\emph{Proceedings of the
  IEEE/CVF International Conference on Computer Vision}}.
  \bibinfo{pages}{10012--10022}.
\newblock


\bibitem[Oksuz et~al\mbox{.}(2021)]%
        {Imbalance}
\bibfield{author}{\bibinfo{person}{Kemal Oksuz}, \bibinfo{person}{Baris~Can
  Cam}, \bibinfo{person}{Sinan Kalkan}, {and} \bibinfo{person}{Emre Akbas}.}
  \bibinfo{year}{2021}\natexlab{}.
\newblock \showarticletitle{Imbalance Problems in Object Detection: {A}
  Review}.
\newblock \bibinfo{journal}{\emph{{IEEE} Trans. Pattern Anal. Mach. Intell.}}
  \bibinfo{volume}{43}, \bibinfo{number}{10} (\bibinfo{year}{2021}),
  \bibinfo{pages}{3388--3415}.
\newblock


\bibitem[Peng et~al\mbox{.}(2020)]%
        {peng2020deep}
\bibfield{author}{\bibinfo{person}{Jizong Peng}, \bibinfo{person}{Guillermo
  Estrada}, \bibinfo{person}{Marco Pedersoli}, {and} \bibinfo{person}{Christian
  Desrosiers}.} \bibinfo{year}{2020}\natexlab{}.
\newblock \showarticletitle{Deep co-training for semi-supervised image
  segmentation}.
\newblock \bibinfo{journal}{\emph{Pattern Recognition}}  \bibinfo{volume}{107}
  (\bibinfo{year}{2020}), \bibinfo{pages}{107269}.
\newblock


\bibitem[Redmon et~al\mbox{.}(2016)]%
        {YOLO}
\bibfield{author}{\bibinfo{person}{Joseph Redmon}, \bibinfo{person}{Santosh
  Divvala}, \bibinfo{person}{Ross Girshick}, {and} \bibinfo{person}{Ali
  Farhadi}.} \bibinfo{year}{2016}\natexlab{}.
\newblock \showarticletitle{You Only Look Once: Unified, Real-Time Object
  Detection}. In \bibinfo{booktitle}{\emph{IEEE Conference on Computer Vision
  and Pattern Recognition}}. \bibinfo{pages}{779--788}.
\newblock


\bibitem[Ren et~al\mbox{.}(2017)]%
        {FasterRCNN}
\bibfield{author}{\bibinfo{person}{Shaoqing Ren}, \bibinfo{person}{Kaiming He},
  \bibinfo{person}{Ross Girshick}, {and} \bibinfo{person}{Jian Sun}.}
  \bibinfo{year}{2017}\natexlab{}.
\newblock \showarticletitle{Faster R-CNN: Towards Real-Time Object Detection
  with Region Proposal Networks}.
\newblock \bibinfo{journal}{\emph{IEEE Transactions on Pattern Analysis and
  Machine Intelligence}} \bibinfo{volume}{39}, \bibinfo{number}{6}
  (\bibinfo{year}{2017}), \bibinfo{pages}{1137--1149}.
\newblock


\bibitem[Russakovsky et~al\mbox{.}(2015)]%
        {russakovsky2015imagenet}
\bibfield{author}{\bibinfo{person}{Olga Russakovsky}, \bibinfo{person}{Jia
  Deng}, \bibinfo{person}{Hao Su}, \bibinfo{person}{Jonathan Krause},
  \bibinfo{person}{Sanjeev Satheesh}, \bibinfo{person}{Sean Ma},
  \bibinfo{person}{Zhiheng Huang}, \bibinfo{person}{Andrej Karpathy},
  \bibinfo{person}{Aditya Khosla}, \bibinfo{person}{Michael Bernstein},
  {et~al\mbox{.}}} \bibinfo{year}{2015}\natexlab{}.
\newblock \showarticletitle{Imagenet large scale visual recognition challenge}.
\newblock \bibinfo{journal}{\emph{International journal of computer vision}}
  \bibinfo{volume}{115}, \bibinfo{number}{3} (\bibinfo{year}{2015}),
  \bibinfo{pages}{211--252}.
\newblock


\bibitem[Shi et~al\mbox{.}(2018)]%
        {shi2018transductive}
\bibfield{author}{\bibinfo{person}{Weiwei Shi}, \bibinfo{person}{Yihong Gong},
  \bibinfo{person}{Chris Ding}, \bibinfo{person}{Zhiheng~MaXiaoyu Tao}, {and}
  \bibinfo{person}{Nanning Zheng}.} \bibinfo{year}{2018}\natexlab{}.
\newblock \showarticletitle{Transductive semi-supervised deep learning using
  min-max features}. In \bibinfo{booktitle}{\emph{Proceedings of the European
  Conference on Computer Vision (ECCV)}}. \bibinfo{pages}{299--315}.
\newblock


\bibitem[Sohn et~al\mbox{.}(2020a)]%
        {sohn2020fixmatch}
\bibfield{author}{\bibinfo{person}{Kihyuk Sohn}, \bibinfo{person}{David
  Berthelot}, \bibinfo{person}{Nicholas Carlini}, \bibinfo{person}{Zizhao
  Zhang}, \bibinfo{person}{Han Zhang}, \bibinfo{person}{Colin~A Raffel},
  \bibinfo{person}{Ekin~Dogus Cubuk}, \bibinfo{person}{Alexey Kurakin}, {and}
  \bibinfo{person}{Chun-Liang Li}.} \bibinfo{year}{2020}\natexlab{a}.
\newblock \showarticletitle{Fixmatch: Simplifying semi-supervised learning with
  consistency and confidence}.
\newblock \bibinfo{journal}{\emph{Advances in Neural Information Processing
  Systems}}  \bibinfo{volume}{33} (\bibinfo{year}{2020}),
  \bibinfo{pages}{596--608}.
\newblock


\bibitem[Sohn et~al\mbox{.}(2020b)]%
        {sohn2020simple}
\bibfield{author}{\bibinfo{person}{Kihyuk Sohn}, \bibinfo{person}{Zizhao
  Zhang}, \bibinfo{person}{Chun-Liang Li}, \bibinfo{person}{Han Zhang},
  \bibinfo{person}{Chen-Yu Lee}, {and} \bibinfo{person}{Tomas Pfister}.}
  \bibinfo{year}{2020}\natexlab{b}.
\newblock \showarticletitle{A simple semi-supervised learning framework for
  object detection}.
\newblock \bibinfo{journal}{\emph{arXiv preprint arXiv:2005.04757}}
  (\bibinfo{year}{2020}).
\newblock


\bibitem[Solovyev et~al\mbox{.}(2021)]%
        {solovyev2021weighted}
\bibfield{author}{\bibinfo{person}{Roman Solovyev}, \bibinfo{person}{Weimin
  Wang}, {and} \bibinfo{person}{Tatiana Gabruseva}.}
  \bibinfo{year}{2021}\natexlab{}.
\newblock \showarticletitle{Weighted boxes fusion: Ensembling boxes from
  different object detection models}.
\newblock \bibinfo{journal}{\emph{Image and Vision Computing}}
  \bibinfo{volume}{107} (\bibinfo{year}{2021}), \bibinfo{pages}{104117}.
\newblock


\bibitem[Song et~al\mbox{.}(2020)]%
        {song2020revisiting}
\bibfield{author}{\bibinfo{person}{Guanglu Song}, \bibinfo{person}{Yu Liu},
  {and} \bibinfo{person}{Xiaogang Wang}.} \bibinfo{year}{2020}\natexlab{}.
\newblock \showarticletitle{Revisiting the sibling head in object detector}. In
  \bibinfo{booktitle}{\emph{Proceedings of the IEEE/CVF Conference on Computer
  Vision and Pattern Recognition}}. \bibinfo{pages}{11563--11572}.
\newblock


\bibitem[Tang et~al\mbox{.}(2021)]%
        {tang2021humble}
\bibfield{author}{\bibinfo{person}{Yihe Tang}, \bibinfo{person}{Weifeng Chen},
  \bibinfo{person}{Yijun Luo}, {and} \bibinfo{person}{Yuting Zhang}.}
  \bibinfo{year}{2021}\natexlab{}.
\newblock \showarticletitle{Humble teachers teach better students for
  semi-supervised object detection}. In \bibinfo{booktitle}{\emph{Proceedings
  of the IEEE/CVF Conference on Computer Vision and Pattern Recognition}}.
  \bibinfo{pages}{3132--3141}.
\newblock


\bibitem[Tarvainen and Valpola(2017)]%
        {tarvainen2017mean}
\bibfield{author}{\bibinfo{person}{Antti Tarvainen} {and}
  \bibinfo{person}{Harri Valpola}.} \bibinfo{year}{2017}\natexlab{}.
\newblock \showarticletitle{Mean teachers are better role models:
  Weight-averaged consistency targets improve semi-supervised deep learning
  results}.
\newblock \bibinfo{journal}{\emph{Advances in neural information processing
  systems}}  \bibinfo{volume}{30} (\bibinfo{year}{2017}).
\newblock


\bibitem[Tian et~al\mbox{.}(2019)]%
        {tian2019fcos}
\bibfield{author}{\bibinfo{person}{Zhi Tian}, \bibinfo{person}{Chunhua Shen},
  \bibinfo{person}{Hao Chen}, {and} \bibinfo{person}{Tong He}.}
  \bibinfo{year}{2019}\natexlab{}.
\newblock \showarticletitle{Fcos: Fully convolutional one-stage object
  detection}. In \bibinfo{booktitle}{\emph{Proceedings of the IEEE/CVF
  international conference on computer vision}}. \bibinfo{pages}{9627--9636}.
\newblock


\bibitem[Vaswani et~al\mbox{.}(2017)]%
        {vaswani2017attention}
\bibfield{author}{\bibinfo{person}{Ashish Vaswani}, \bibinfo{person}{Noam
  Shazeer}, \bibinfo{person}{Niki Parmar}, \bibinfo{person}{Jakob Uszkoreit},
  \bibinfo{person}{Llion Jones}, \bibinfo{person}{Aidan~N Gomez},
  \bibinfo{person}{{\L}ukasz Kaiser}, {and} \bibinfo{person}{Illia
  Polosukhin}.} \bibinfo{year}{2017}\natexlab{}.
\newblock \showarticletitle{Attention is all you need}.
\newblock \bibinfo{journal}{\emph{Advances in neural information processing
  systems}}  \bibinfo{volume}{30} (\bibinfo{year}{2017}).
\newblock


\bibitem[Wang et~al\mbox{.}(2021)]%
        {wang2021combating}
\bibfield{author}{\bibinfo{person}{Zhenyu Wang}, \bibinfo{person}{Ya-Li Li},
  \bibinfo{person}{Ye Guo}, {and} \bibinfo{person}{Shengjin Wang}.}
  \bibinfo{year}{2021}\natexlab{}.
\newblock \showarticletitle{Combating Noise: Semi-supervised Learning by Region
  Uncertainty Quantification}.
\newblock \bibinfo{journal}{\emph{Advances in Neural Information Processing
  Systems}}  \bibinfo{volume}{34} (\bibinfo{year}{2021}).
\newblock


\bibitem[Xie et~al\mbox{.}(2020a)]%
        {xie2020unsupervised}
\bibfield{author}{\bibinfo{person}{Qizhe Xie}, \bibinfo{person}{Zihang Dai},
  \bibinfo{person}{Eduard Hovy}, \bibinfo{person}{Thang Luong}, {and}
  \bibinfo{person}{Quoc Le}.} \bibinfo{year}{2020}\natexlab{a}.
\newblock \showarticletitle{Unsupervised data augmentation for consistency
  training}.
\newblock \bibinfo{journal}{\emph{Advances in Neural Information Processing
  Systems}}  \bibinfo{volume}{33} (\bibinfo{year}{2020}),
  \bibinfo{pages}{6256--6268}.
\newblock


\bibitem[Xie et~al\mbox{.}(2020b)]%
        {xie2020self}
\bibfield{author}{\bibinfo{person}{Qizhe Xie}, \bibinfo{person}{Minh-Thang
  Luong}, \bibinfo{person}{Eduard Hovy}, {and} \bibinfo{person}{Quoc~V Le}.}
  \bibinfo{year}{2020}\natexlab{b}.
\newblock \showarticletitle{Self-training with noisy student improves imagenet
  classification}. In \bibinfo{booktitle}{\emph{Proceedings of the IEEE/CVF
  conference on computer vision and pattern recognition}}.
  \bibinfo{pages}{10687--10698}.
\newblock


\bibitem[Xu et~al\mbox{.}(2021)]%
        {xu2021end}
\bibfield{author}{\bibinfo{person}{Mengde Xu}, \bibinfo{person}{Zheng Zhang},
  \bibinfo{person}{Han Hu}, \bibinfo{person}{Jianfeng Wang},
  \bibinfo{person}{Lijuan Wang}, \bibinfo{person}{Fangyun Wei},
  \bibinfo{person}{Xiang Bai}, {and} \bibinfo{person}{Zicheng Liu}.}
  \bibinfo{year}{2021}\natexlab{}.
\newblock \showarticletitle{End-to-end semi-supervised object detection with
  soft teacher}. In \bibinfo{booktitle}{\emph{Proceedings of the IEEE/CVF
  International Conference on Computer Vision}}. \bibinfo{pages}{3060--3069}.
\newblock


\bibitem[Yang et~al\mbox{.}(2021)]%
        {yang2021interactive}
\bibfield{author}{\bibinfo{person}{Qize Yang}, \bibinfo{person}{Xihan Wei},
  \bibinfo{person}{Biao Wang}, \bibinfo{person}{Xian-Sheng Hua}, {and}
  \bibinfo{person}{Lei Zhang}.} \bibinfo{year}{2021}\natexlab{}.
\newblock \showarticletitle{Interactive self-training with mean teachers for
  semi-supervised object detection}. In \bibinfo{booktitle}{\emph{Proceedings
  of the IEEE/CVF Conference on Computer Vision and Pattern Recognition}}.
  \bibinfo{pages}{5941--5950}.
\newblock


\bibitem[Zhang et~al\mbox{.}(2021)]%
        {zhang2021flexmatch}
\bibfield{author}{\bibinfo{person}{Bowen Zhang}, \bibinfo{person}{Yidong Wang},
  \bibinfo{person}{Wenxin Hou}, \bibinfo{person}{Hao Wu},
  \bibinfo{person}{Jindong Wang}, \bibinfo{person}{Manabu Okumura}, {and}
  \bibinfo{person}{Takahiro Shinozaki}.} \bibinfo{year}{2021}\natexlab{}.
\newblock \showarticletitle{Flexmatch: Boosting semi-supervised learning with
  curriculum pseudo labeling}.
\newblock \bibinfo{journal}{\emph{Advances in Neural Information Processing
  Systems}}  \bibinfo{volume}{34} (\bibinfo{year}{2021}),
  \bibinfo{pages}{18408--18419}.
\newblock


\bibitem[Zhang et~al\mbox{.}(2018)]%
        {zhang2018mixup}
\bibfield{author}{\bibinfo{person}{Hongyi Zhang}, \bibinfo{person}{Moustapha
  Cisse}, \bibinfo{person}{Yann~N Dauphin}, {and} \bibinfo{person}{David
  Lopez-Paz}.} \bibinfo{year}{2018}\natexlab{}.
\newblock \showarticletitle{mixup: Beyond Empirical Risk Minimization}. In
  \bibinfo{booktitle}{\emph{International Conference on Learning
  Representations}}.
\newblock


\bibitem[Zhou et~al\mbox{.}(2021)]%
        {zhou2021instant}
\bibfield{author}{\bibinfo{person}{Qiang Zhou}, \bibinfo{person}{Chaohui Yu},
  \bibinfo{person}{Zhibin Wang}, \bibinfo{person}{Qi Qian}, {and}
  \bibinfo{person}{Hao Li}.} \bibinfo{year}{2021}\natexlab{}.
\newblock \showarticletitle{Instant-teaching: An end-to-end semi-supervised
  object detection framework}. In \bibinfo{booktitle}{\emph{Proceedings of the
  IEEE/CVF Conference on Computer Vision and Pattern Recognition}}.
  \bibinfo{pages}{4081--4090}.
\newblock


\bibitem[Zhu et~al\mbox{.}(2020)]%
        {zhu2020deformable}
\bibfield{author}{\bibinfo{person}{Xizhou Zhu}, \bibinfo{person}{Weijie Su},
  \bibinfo{person}{Lewei Lu}, \bibinfo{person}{Bin Li},
  \bibinfo{person}{Xiaogang Wang}, {and} \bibinfo{person}{Jifeng Dai}.}
  \bibinfo{year}{2020}\natexlab{}.
\newblock \showarticletitle{Deformable DETR: Deformable Transformers for
  End-to-End Object Detection}. In \bibinfo{booktitle}{\emph{International
  Conference on Learning Representations}}.
\newblock


\end{thebibliography}










\end{document}